\documentclass[11pt]{article}

\usepackage{acl}
\usepackage{times}
\usepackage{latexsym}
\usepackage{marvosym}
\usepackage[T1]{fontenc}
\usepackage[utf8]{inputenc}
\usepackage{microtype}
\usepackage{courier}

\usepackage{graphicx}
\usepackage{booktabs}
\usepackage{multirow}
\usepackage{array}
\usepackage{amsmath,amssymb,amsthm,bm}
\usepackage{enumitem}
\usepackage{xcolor}
\definecolor{aclpubdarkblue}{HTML}{000099}
\usepackage[most]{tcolorbox}
\usepackage{listings}
\usepackage{url}
\usepackage{multicol}
\usepackage{placeins}
\usepackage{needspace}
\tcbuselibrary{listings,breakable,skins}
\lstdefinestyle{promptstyle}{
  basicstyle=\ttfamily\scriptsize,
  breaklines=true,
  breakindent=0pt,
  columns=fullflexible,
  keepspaces=true,
  numbers=left,
  numberstyle=\tiny\color{gray},
  numbersep=8pt,
  showstringspaces=false,
  upquote=true,
}
\newtcblisting{promptbox}[1]{
  breakable, enhanced, listing only,
  listing options={style=promptstyle},
  colback=gray!7, colframe=gray!55, boxrule=0.4pt,
  arc=1pt, left=2.2em, top=2pt, bottom=2pt,
  fonttitle=\bfseries\footnotesize, coltitle=black,
  colbacktitle=gray!18, title={#1}}

\hypersetup{citecolor=aclpubdarkblue, linkcolor=aclpubdarkblue, urlcolor=aclpubdarkblue}

\newcommand{\spire}{\textbf{\textsc{SPIRE}}}
\newcommand{\Q}{\mathcal{Q}}
\newcommand{\C}{\mathcal{C}}
\newcommand{\E}{\mathcal{E}}
\newcommand{\textunit}{u}

\newtheoremstyle{spdef}{6pt}{6pt}{}{}{\bfseries}{.}{ }{}
\theoremstyle{spdef}
\newtheorem{spdefn}{Definition}

\title{Extending AI for Research to the Humanities: \\
A Multi-Agent Framework for Evidence-Grounded Scholarship}

\author{
  \textbf{Yating Pan}\textsuperscript{1,2} \quad
  \textbf{Jiajun Zhang}\textsuperscript{1,2} \quad
  \textbf{Jun Wang}\textsuperscript{1,2,4,*,\Letter} \quad
  \textbf{Qi Su}\textsuperscript{2,3,4,\ensuremath{\dagger}} \\
  \textsuperscript{1}Department of Information Management, Peking University \\
  \textsuperscript{2}Research Center for Digital Humanities, Peking University \\
  \textsuperscript{3}School of Foreign Languages, Peking University \\
  \textsuperscript{4}Institute for Artificial Intelligence, Peking University, Beijing, China \\
  {\small\texttt{ytpan25@stu.pku.edu.cn} \quad
  \texttt{2601111285@stu.pku.edu.cn}} \\
  {\small\texttt{junwang@pku.edu.cn} \quad
  \texttt{sukia@pku.edu.cn}}
}

\begin{document}
\maketitle
\begingroup
\renewcommand{\thefootnote}{\fnsymbol{footnote}}
\footnotetext[1]{Lead corresponding author.}
\footnotetext[2]{Corresponding author.}
\endgroup

\begin{abstract}
LLM-based research agents have advanced rapidly in science and engineering, where research is organized around executable experiments, code, and quantitative signals. Humanities scholarship, however, requires interpretive, evidence-grounded arguments over primary sources, whose value rests on faithful quotation, verifiable provenance, and deep interpretation. Existing research agents use general planning, tool use, and reflection, leaving these scholarly operations implicit. To address this gap, we introduce \spire{} (\textbf{S}cholarly-\textbf{P}rimitives-\textbf{I}nspired \textbf{R}esearch \textbf{E}ngine), a multi-agent framework that realizes Scholarly Primitives, a typology of basic humanities scholarship practices, as cooperating roles over a multi-scale close-reading substrate of passages, intra-context graph communities, and cross-context semantic clusters. On a benchmark of peer-reviewed papers in classical Chinese and Greco-Roman Latin scholarship, \spire{} recovers substantially more cited primary-source evidence and receives higher blind human and LLM ratings on four scholarly quality dimensions than LLM, RAG, and generic agentic baselines. Retrieval-tier and agent ablations, with retrieval-volume and answer-length controls, trace its advantage to targeted retrieval, structured evidence selection, and claim-evidence binding. Code and benchmark data are released at \url{https://github.com/YatingPan/SPIRE}.
\end{abstract}

\begin{figure*}[!t]
\centering
\includegraphics[width=\textwidth]{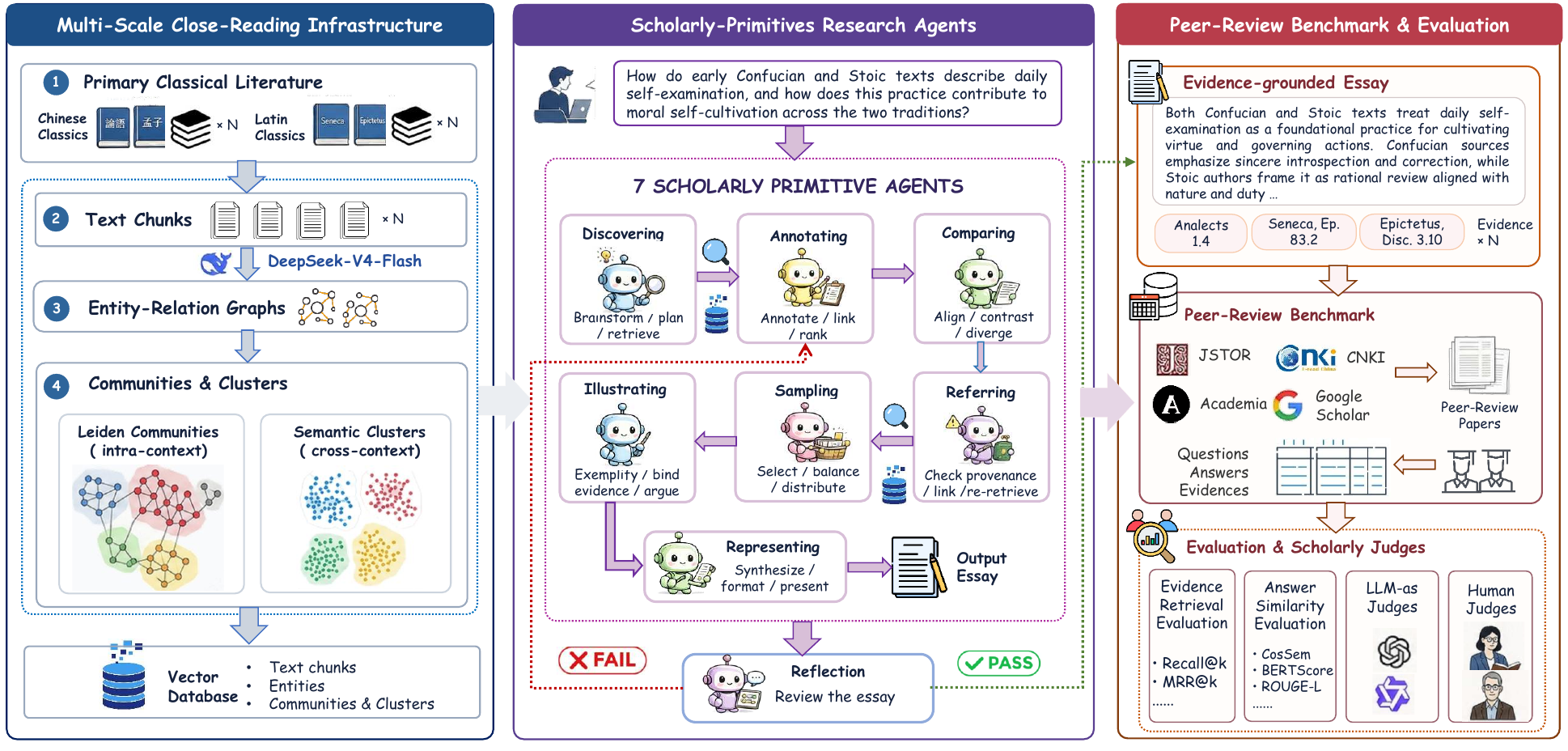}
\caption{Overview of \spire{}: compiles classics into a multi-scale close-reading store, answers a research question with seven scholarly-primitive agents over an EvidencePool, and evaluates on a peer-reviewed-paper benchmark.}
\label{fig:overview}
\end{figure*}

\section{Introduction}
\label{sec:intro}

LLM-based agents have recently achieved strong performance in science and engineering. In fields such as chemistry and materials science, they design experiments, analyze data, and test hypotheses \citep{boiko2023coscientist,ghafarollahi2024sciagents,chen2025scienceagentbench}. In computer science and software engineering, they survey literature, write and test code, and plan experiments \citep{lu2024aiscientist,romera2024funsearch,schmidgall2025agentlab,he2025pasa,jimenez2024swebench,zhang2025aflow,zhuge2024gptswarm}. They also conduct quantitative analysis and simulate participants or populations in the social sciences \citep{park2023generative,horton2023homo,argyle2023out,li2023econagent,piao2025agentsociety,gao2024llmabmsurvey}. Advances in workflow orchestration \citep{wu2023autogen}, multi-agent collaboration \citep{chen2024ioa}, retrieval-augmented generation \citep{lewis2020rag,gao2023ragsurvey}, reasoning and acting loops \citep{yao2023react}, self-reflection \citep{shinn2023reflexion}, and reinforcement learning \citep{deepseekr1} enable these systems to coordinate complex research tasks. Across these domains, research is typically organized around executable procedures, verifiable numerical outcomes, code, or data analysis.

This mode of agentic assistance remains less developed in text-centered humanities scholarship, whose core practices require different capabilities.\footnote{By humanities research, we focus on text-centered scholarship in history, philosophy, literature, philology, and so on, excluding computationally oriented areas such as quantitative history and digital humanities \citep{moretti2013distant,jockers2013macroanalysis,underwood2019distant,rockwell2016hermeneutica}.} Humanities scholarship is qualitative, interpretive, and evidence-grounded. Scholarly training emphasizes selecting and closely reading primary sources, mining and weighing evidence for and against a claim, and developing original interpretations. Each interpretation rests on faithful quotation, verifiable provenance, and primary-source evidence \citep{smith2016closereading}. Compared with fixed experimental procedures or reproducible code and computation, humanistic inquiry places greater weight on evidential sufficiency and interpretive depth. Multiple well-supported readings may remain possible, leaving outcomes open-ended and difficult to verify with a single quantitative signal. Agentic assistance must therefore make scholarly operations and evidence paths explicit.

Although humanities scholarship is interpretive and open-ended, scholarly training has long been organized around recurring practices that offer an explicit basis for agentic workflows. Scholarly Primitives theory characterizes the basic activities shared across humanities scholarship as discovering, annotating, comparing, referring, sampling, illustrating, and representing \citep{unsworth2000scholarly}. Building on this perspective, we operationalize these practices as an executable multi-agent workflow. \spire{} (\textbf{S}cholarly-\textbf{P}rimitives-\textbf{I}nspired \textbf{R}esearch \textbf{E}ngine) coordinates agents over a typed primary-source EvidencePool to discover sources, annotate evidence, compare perspectives, verify provenance, sample passages, bind citations, and synthesize a citation-grounded argument. To our knowledge, \spire{} is among the first multi-agent research frameworks designed explicitly for evidence-grounded interpretive scholarship and models a complete path from source discovery to argumentative synthesis.

The EvidencePool in turn requires a retrieval substrate that reflects how humanists engage with primary texts. Close reading grounds interpretation and argument in the detailed selection and examination of primary sources. RAG and its variants improve semantic access by embedding passages and, increasingly, graphs or structured knowledge \citep{lewis2020rag,gao2023ragsurvey,edge2025graphrag,luo2025hypergraphrag,han2025graphragsurvey}. Humanistic reading also situates passages across multiple contextual scales, including explicit comparison with related works \citep{kristeva1980intertext} and recurring conceptual formations across texts \citep{lovejoy1936great}. We therefore build a multi-scale close-reading substrate for the scholarly-primitive agents: passage embeddings for direct text evidence, entity-relation graph communities for intra-context relational evidence, and semantic clusters for cross-context recurrent evidence.

Evaluating such assistance requires measures that reflect both evidence grounding and interpretive essay quality. To this end, we construct a peer-reviewed-paper benchmark over classical Chinese and Greco-Roman Latin scholarship, extracting each paper's research question, main findings, and cited primary-source evidence. Given only the title and research question, a system must perform the scholarly steps and produce an evidence-grounded essay. We evaluate two complementary dimensions: cited-evidence recovery at the sentence, section, and work levels, testing whether the system retrieves primary sources used in the reference paper; and blind scholarly evaluation by both expert human reviewers and LLM judges assessing answer accuracy, argument depth, coverage, and evidence quality. Compared with Naive LLM, Text RAG, GraphRAG, and generic agentic baselines, \spire{} attains $44.3\%$ evidence recall versus $30.0\%$ for the strongest baseline and receives higher judge scores on all four aspects. Dynamic and frozen-pool ablations separate iterative evidence acquisition from post-retrieval curation. Retrieval-volume and answer-length controls test the advantage under fixed volume and comparable response length.

Our contributions are as follows:
\begin{enumerate}[leftmargin=1.4em,itemsep=2pt]
\item We introduce \spire{}, a multi-agent framework for evidence-grounded humanities scholarship that operationalizes Scholarly Primitives as coordinated agent workflows for interpretive research over primary sources.

\item We propose a multi-scale close-reading substrate combining passages, graph communities, and semantic clusters to support direct evidence grounding, contextual interpretation, and recurrent conceptual analysis.

\item We construct a peer-reviewed-paper benchmark of 406 papers centered on primary-source evidence and scholarly judging, providing a reusable testbed for future humanities research agents.
\end{enumerate}

\section{Background and Related Work}
\label{sec:related}

\subsection{LLM agents for research and their evaluation}
\label{sec:rw-agents}

LLM agents have emerged as a major paradigm for AI-assisted research \citep{zheng2025automation}, especially in domains with explicit operational or verification signals. In the natural sciences, agents connect LLMs with laboratory APIs and domain-specific tools to plan and execute experiments \citep{boiko2023coscientist,bran2024chemcrow}. In computer science, they automate parts of the research cycle, from ideation to code execution and paper writing \citep{lu2024aiscientist,li2024mlrcopilot,schmidgall2025agentlab}. In quantitative social science, they simulate individuals, populations, and economic activity \citep{park2023generative,li2023econagent,piao2025agentsociety}. Agent performance is commonly evaluated through task completion and executable outcomes. For example, ScienceAgentBench evaluates generated programs and execution results on data-driven scientific tasks \citep{chen2025scienceagentbench}, while SWE-bench tests whether agents resolve software issues through repository tests \citep{jimenez2024swebench}.

Research agents, however, remain less common in the humanities. Humanistic inquiry centers on evidence-based interpretation of primary-source texts and may support multiple defensible interpretations. This limits the fit of agents and benchmarks built around executable tasks or reproducible quantitative results. \spire{} addresses the gap by modeling established scholarly practices as an agent workflow and evaluating primary-source evidence retrieval and essay quality against domain-specific criteria.

\subsection{Retrieval-augmented generation for research}
\label{sec:rw-rag}

In addition to research agents, retrieval-augmented generation (RAG) supports AI-assisted research through literature reading, evidence synthesis, and knowledge retrieval. It grounds model outputs in retrieved passages to reduce unsupported claims and provide citation-backed answers \citep{lewis2020rag}. PaperQA retrieves and assesses passages across full-text scientific articles \citep{lala2023paperqa}, while OpenScholar searches millions of papers and synthesizes citation-backed responses through retrieval and self-feedback \citep{asai2024openscholar}. Graph-based variants extract structured relational knowledge from documents to strengthen global retrieval. GraphRAG uses entity relations and community summaries, while HyperGraphRAG represents higher-order relations through hyperedges \citep{edge2025graphrag,luo2025hypergraphrag}. Such systems have supported knowledge-base question answering in medical and technical domains \citep{wu2025medicalgraphrag,kostadinov2025medsumgraph,padhi2025documentgraphrag}.

In the humanities, RAG has mainly supported access to source collections. FolkRAG retrieves cultural heritage materials, while Topic-RAG uses topic-guided retrieval over historical newspapers \citep{xie2025folkrag,murugaraj2025topicrag}. These systems improve information access. Direct research assistance must also organize and interpret textual evidence across multiple contexts. \spire{} addresses this need through a multi-scale close-reading substrate of passages, graph communities, and semantic clusters.

\subsection{Scholarly Primitives and close-reading theory}
\label{sec:rw-theory}

Humanities research paradigms provide the domain structure needed to adapt the general agents and retrieval methods above. Experimental science is often organized into repeatable, fine-grained procedures. Humanistic inquiry follows more flexible paths, yet its long-established practices can still be summarized as shared scholarly operations. Scholarly Primitives identifies these operations as discovering, annotating, comparing, referring, sampling, illustrating, and representing \citep{unsworth2000scholarly}. Work on scholarly information practices and humanities visualization similarly describes humanistic research through information seeking, interpretation, organization, and representation \citep{palmer2009scholarly,drucker2011humanities}. We use this framework because it turns broad scholarly practices into explicit roles that agents can perform.

Agent roles define the research workflow. Primary-source interpretation also requires a retrieval substrate that situates evidence within textual, historical, and conceptual contexts. Close-reading traditions offer two complementary perspectives. Contextual reading, associated with Skinner and the Cambridge School, links meaning to historically attested texts, authors, intentions, and arguments \citep{skinner1969meaning}. Recurrent reading, drawing on Lovejoy's history of ideas, traces conceptual formations across periods, works, and traditions \citep{lovejoy1936great}. It can surface relevant evidence when no direct textual link is documented. Together, these modes organize evidence within and across contexts. \spire{} maps them to multi-scale retrieval: passages for direct quotation, graph communities for intra-context relational evidence, and semantic clusters for cross-context recurrent evidence.

\section{Methods: \spire{}}
\label{sec:method}

\subsection{Overview}
\label{sec:problem}

Building on the scholarly primitives and close-reading modes of \S\ref{sec:rw-theory}, \spire{} combines a multi-scale evidence store with seven cooperating research agents. A user supplies a classical-text corpus, an entity/relation schema, and a research question. The system returns an evidence-grounded essay with traceable primary-source citations (Figure~\ref{fig:overview}). Let \(\C\) be a corpus segmented into text units \(\textunit\), and let \(\Q\) be the research question. \spire{} produces a ranked EvidencePool \(\E(\Q)\) and an essay \(A(\Q)\) grounded in that pool.

The system has two connected components. The multi-scale close-reading infrastructure (\S\ref{sec:retrieval}) organizes \(\C\) as passages, an entity-relation graph, intra-context communities, and cross-context clusters. The scholarly-primitive workflow (\S\ref{sec:primitives}) turns a question into sub-questions, retrieves evidence from these levels, and passes a shared EvidencePool through seven agents. The agents interpret, compare, verify, and select evidence before synthesizing the essay. Reflection reviews the result and can request targeted evidence for up to two rounds.

\subsection{Scholarly-primitive agents and the EvidencePool}
\label{sec:primitives}

\spire{} realizes the seven Scholarly Primitives (\S\ref{sec:rw-theory}) as cooperating agent roles. They share an evolving EvidencePool that connects retrieval, evidence analysis, and writing. Table~\ref{tab:primitives} summarizes their operations, while Figure~\ref{fig:workflow} follows one complete run.

\subsubsection{Agent roles and the shared EvidencePool}

\begin{table}[!ht]
\centering\scriptsize
\setlength{\tabcolsep}{1.5pt}
\renewcommand{\arraystretch}{0.92}
\begin{tabular}{@{}clp{2.55cm}p{2.55cm}@{}}
\toprule
 & Primitive & Main operation & Effect on pool or output \\
\midrule
$\mathcal{D}$ & Discovering & Plan sub-questions and search three tiers. & Initialize the EvidencePool. \\
$\mathcal{N}$ & Annotating & Interpret and score retrieved passages. & Annotate and re-rank evidence. \\
$\mathcal{P}$ & Comparing & Compare question-defined source groups. & Add parallels and divergences. \\
$\mathcal{R}$ & Referring & Check provenance and source links. & Validate and expand the pool. \\
$\mathcal{S}$ & Sampling & Balance question, work, and tradition coverage. & Select the evidence subset. \\
$\mathcal{I}$ & Illustrating & Exemplify claims with retrieved passages. & Bind claims to evidence IDs. \\
$\mathcal{T}$ & Representing & Organize evidence-bound claims. & Produce the final essay. \\
\bottomrule
\end{tabular}
\caption{The seven scholarly primitives and their effects on the shared EvidencePool and research output.}
\label{tab:primitives}
\end{table}

\begin{spdefn}[EvidencePool]
\label{def:evpool}
The \emph{EvidencePool} \(\E_k=[e_1,\ldots,e_n]\) is the shared research workspace after workflow stage \(k\). Each item \(e_j=(\textunit_j,m_j,r_j,z_j)\) contains a primary-source text unit \(\textunit_j\), source metadata \(m_j\), retrieval provenance \(r_j\), and agent-added information \(z_j\). The metadata identifies its author, work, chapter or section, language, and canonical locator. Retrieval provenance records the sub-question and retrieval tier that produced the item. Agent-added information can include an interpretation, relevance score, comparison relation, provenance check, or citation binding.
\end{spdefn}

Discovering initializes \(\E_0\). Later agents can add evidence, attach interpretations, revise its ranking, organize comparative relations, verify sources, and select a final subset. The primary text and its locator remain linked throughout the workflow. Agents can therefore quote a passage directly and read it within its section and work context.

\subsubsection{Query planning and evidence acquisition}
\label{sec:acquisition}

Given \(\Q\), Discovering (\(\mathcal{D}\)) produces a query plan
\(\pi(\Q)=(\{q_i\},\mathrm{lang},c,\mathrm{anchors},\mathrm{hints})\).
The plan identifies source languages, literal work and concept anchors, and the comparison scope \(c\). It also decomposes the question into two to four sub-questions \(q_i\), each tagged by granularity and analytical perspective. Retrieval hints supply related concepts, native terms, and candidate works that broaden the search.

Each \(q_i\) queries three tools over the multi-scale store (\S\ref{sec:retrieval}). The passage tier \(L\) retrieves local text units and their entities. The community tier \(G\) retrieves intra-context reports and expands them to supporting units. The cluster tier \(I\) retrieves cross-context semantic affinities and follows them to their supporting entities and passages. Their ranked union initializes the EvidencePool:
\begin{equation}
\E_0(\Q)=
\operatorname{Rank}\!\left(
\bigcup_i\bigcup_{t\in\{L,G,I\}}T_t(q_i)
\right).
\label{eq:fusion}
\end{equation}
The ranking combines explicit work and anchor matches with semantic similarity. Every result retains its source locator, retrieval tier, and originating sub-question. Broad community and cluster matches can therefore be traced back to citable primary-source passages.

\subsubsection{Orchestration, synthesis, and reflection}
\label{sec:orchestration}

After Discovering creates \(\E_0\), the orchestrator runs
\[
\mathcal{D}\!\to\!\mathcal{N}\!\to\!\mathcal{P}\!\to\!
\mathcal{R}\!\to\!\mathcal{S}\!\to\!\mathcal{I}\!\to\!
\mathcal{T}\!\to\!\mathrm{Reflection}.
\]
Annotating relates each passage to a sub-question and assigns interpretive relevance. Comparing organizes the evidence groups specified by the query plan, such as named works, authors, or traditions. It records parallels, divergences, unilateral concepts, and inverted emphases. Referring checks work and section provenance, follows explicit source connections, and retrieves evidence for missing named sources. Sampling then selects a balanced subset across sub-questions, works, and traditions.

Illustrating forms evidence-grounded claims and binds them to stable evidence IDs. Representing organizes these claims into the final essay. A separate Reflection step reviews question coverage, evidence balance, identifiable gaps, and source support. A diagnosed gap produces a targeted retrieval query and returns the new evidence to annotation and the downstream agents. The process ends when the review passes or reaches two rounds.

\begin{figure}[!ht]
\centering
\makebox[\columnwidth][r]{\includegraphics[width=1.05\columnwidth]{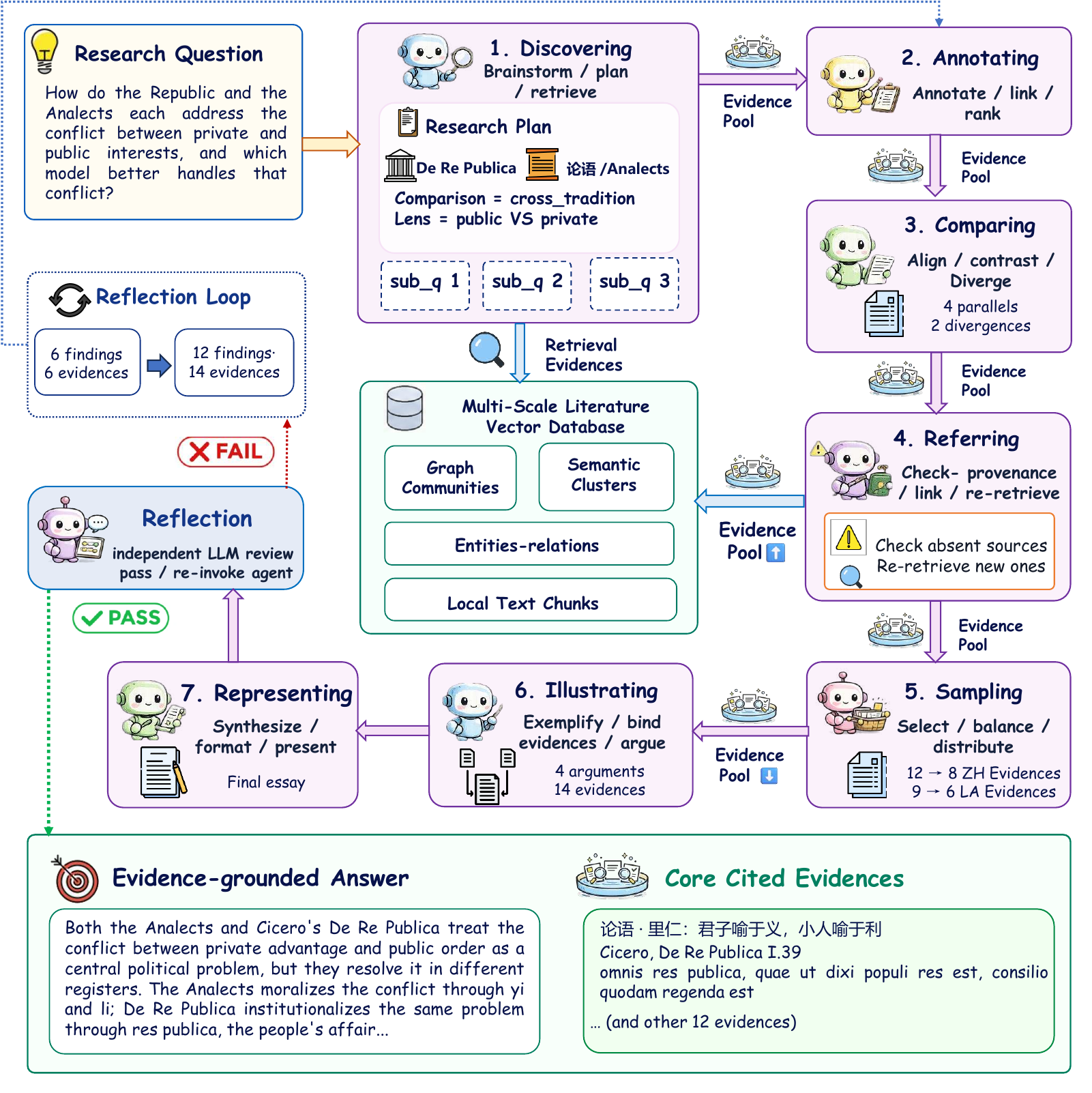}}
\caption{A cross-tradition run on Cicero's \emph{De Re Publica} and the \emph{Analects}. Discovering forms the query plan and initial EvidencePool. Later agents annotate, compare, validate, select, and synthesize its evidence; Reflection can trigger targeted retrieval.}
\label{fig:workflow}
\end{figure}

Figure~\ref{fig:workflow} illustrates the sequence for a comparison between Cicero's \emph{De Re Publica} and the \emph{Analects}. These two works define the comparison groups. Discovering creates three sub-questions and retrieves from all three tiers. Sampling retains eight Chinese and six Latin evidence items, and Illustrating binds four arguments to these fourteen items before Representing writes the essay. Per-agent prompt contracts and output schemas appear in App.~\ref{app:agentprompts}.

\subsection{Multi-scale close-reading infrastructure}
\label{sec:retrieval}

\subsubsection{Corpora and text units}

We instantiate \spire{} on two classical corpora central to humanities scholarship: classical Chinese (\emph{Zhongguo Xueshu Mingzhu Tiyao} \citep{zhongguo1992mingzhu}) and Greco-Roman Latin (\emph{The Latin Library} \citep{latinlibrary}). History graduate students annotate every work with metadata including author, date, and alternative titles. Both span antiquity to the modern period and are standard primary sources in history, philology, and literary study. Table~\ref{tab:corpus} reports their sizes and time spans.

We segment each corpus with \texttt{tiktoken} into \(1000\)-token text units with \(100\)-token overlap, giving \(U=\{u_1,\ldots,u_M\}\). The unit size preserves the surrounding argument, speaker, and citation context needed for classical passages. Overlap reduces fragmentation at unit boundaries. Each unit remains linked to its work and section metadata.

\begin{table}[t]
\centering\scriptsize
\setlength{\tabcolsep}{2pt}
\begin{tabular}{@{}lrrr@{}}
\toprule
 & Classical Chinese & Classical Latin & Total \\
\midrule
Documents & 226 & 710 & 936 \\
Text Chunks & 51,726 & 35,117 & 86,843 \\
Entities & 133,686 & 75,318 & 209,004 \\
Relation edges & 286,123 & 176,405 & 462,528 \\
Leiden communities & 7,452 & 4,647 & 12,099 \\
HDBSCAN clusters & 9,795 & 6,480 & 16,275 \\
Time span & 8c BCE--19c CE & 5c BCE--20c CE & -- \\
\bottomrule
\end{tabular}
\caption{Classical sources corpus and retrieval substrate}
\label{tab:corpus}
\vspace{-12pt}
\end{table}

\subsubsection{Building the multi-scale store}

A user defines an entity/relation schema, and an LLM extracts a typed graph from the text units. Our exemplar is \emph{conceptual history}, a task shared across history, literary studies, and philology. The schema contains three entity types, \emph{Concept}, \emph{Person}, and \emph{Work}, together with free, text-grounded relations. Its few-shot prompt contains one Chinese and one Latin example (App.~\ref{app:prompts}). \textsc{DeepSeek-V4-Flash} performs extraction. Format validation, deduplication, and cross-unit entity merging yield a typed graph \(\mathcal{G}=(\mathcal{V},\mathcal{R})\), where \(\mathcal{R}\subseteq\mathcal{V}\times\mathcal{V}\) contains directed semantic relations. Per-tradition counts appear in Table~\ref{tab:corpus}, and schemas for other fields appear in App.~\ref{app:examples}.

Two higher levels organize the graph at different contextual scales. The \textbf{intra-context} level captures corpus-attested relations among works, persons, and concepts, following contextual reading \citep{skinner1969meaning}. Leiden community detection \citep{traag2019leiden} groups these connected entities. The \textbf{cross-context} level captures recurrent conceptual affinity among semantically related entities, following history-of-ideas research \citep{lovejoy1936great,betti2016computational}. HDBSCAN \citep{mcinnes2017hdbscan} clusters their description embeddings. \textsc{DeepSeek-V4-Flash} names and summarizes every community and cluster from its leading entities. Both levels are built separately for each tradition at four fine-to-coarse resolutions, producing parallel views of the corpus (App.~Table~\ref{tab:artifacts}).

\subsubsection{Unified indexing and provenance}

A normalized relational schema connects every retrieval level to the primary texts. Each community or cluster links to its member entities and relations. Each entity and relation links to supporting text units, while every text unit records its work, chapter, section, and canonical citation. This creates a continuous provenance path from a high-level report to citable source passages.

Graph extraction and community or cluster reports use \textsc{DeepSeek-V4-Flash} with thinking enabled and temperature \(0.1\) (App.~\ref{app:prompts}). The answer-generation agents use the generator setting in \S\ref{sec:baselines}. One multilingual BGE-M3 encoder \citep{chen2024bge} (\(d{=}1024\), dense) embeds text units, entities, reports, and queries. For a query \(q\) and indexed object \(x\), retrieval uses the inner product
\(\mathrm{sim}(q,x)=\phi(q)^{\!\top}\phi(x)\).
The store records each result's tier \(t\), work match \(s_w\), anchor match \(s_a\), and cosine score \(s_{\cos}\). These signals support the ranking in Equation~\ref{eq:fusion}, while the tier and source links preserve retrieval provenance. All objects are stored in DuckDB with FAISS inner-product indexes \citep{johnson2019faiss}.

\section{Experiments}
\label{sec:experiments}

\subsection{Benchmark construction}
\label{sec:benchmark}

To evaluate AI-assisted humanities research, we build a peer-reviewed-paper benchmark in which each paper supplies a research question, a humanist answer, and the primary-source evidence supporting that answer. We collect $406$ refereed papers in English or Chinese on classical Chinese and Latin texts from JSTOR, Academia, Google Scholar, and CNKI (Table~\ref{tab:dataset}). Two master's students in history and philology read every paper and manually extract its research question, abstract, findings, and cited primary-source evidence (example in App.~\ref{app:examples}). The records preserve the paper's wording and finest available locator. Annotators transcribe quoted passages verbatim and record the work, book, chapter, section, or line when the paper supplies a locator. This process yields three evidence granularities: quoted sentences, cited sections, and cited works. The latter two also capture indirect citation through summary, paraphrase, or allusion, motivating the retrieval metrics in Table~\ref{tab:auto}.

Manual annotation captures the flexible structure of humanities articles and their visual evidence cues, including indentation, regular-script Chinese, and italicized Latin. Two annotators independently re-annotated a stratified sample of $100$ papers. Question and finding agreement is $0.86$, source-work Set F1 is $0.94$, normalized-locator agreement is $0.89$, and quotation span-overlap F1 is $0.92$; a domain expert adjudicated disagreements. At evaluation time, each system receives the paper title and research question and returns an evidence-grounded essay. We assess its direct and indirect evidence use together with its overall scholarly quality.

\begin{table}[t]
\centering\small
\setlength{\tabcolsep}{5pt}
\begin{tabular}{@{}lrrrr@{}}
\toprule
Category & Papers & EN/ZH & mean Ev. & mean Pg. \\
\midrule
ZH, single & 103 & 56/47 & 28.8 & 24 \\
ZH, multi  & 61  & 49/12 & 28.3 & 26 \\
LA, single & 89  & 55/34 & 24.3 & 39 \\
LA, multi  & 60  & 51/9  & 30.8 & 36 \\
Cross      & 93  & 75/18 & 32.7 & 33 \\
\midrule
Total      & 406 & 286/120 & 28.9 & 27.8 \\
\bottomrule
\end{tabular}
\caption{Benchmark statistics ($n{=}406$). ZH/LA = Chinese/Latin
tradition; \emph{single}/\emph{multi} = one/several classics within a tradition,
\emph{Cross} = Chinese--Latin comparison; EN/ZH = paper-language;
mean Ev./Pg.\ = mean cited primary-source evidences / page counts. The five
categories are the strata for the $100$-paper judge study (\S\ref{sec:metrics}).}
\label{tab:dataset}
\vspace{-12pt}
\end{table}

\begin{table*}[!t]\centering\scriptsize\setlength{\tabcolsep}{2pt}\renewcommand{\arraystretch}{0.96}
\begin{tabular*}{\textwidth}{@{\extracolsep{\fill}}l rrrrrr rrr@{}}
\toprule
& \multicolumn{6}{c}{Evidence retrieval} & \multicolumn{3}{c}{Answer similarity} \\
\cmidrule(lr){2-7}\cmidrule(lr){8-10}
System & EvidenceR & EvidenceR@10 & WorkR@10 & SentenceR@10 & SectionR@10 & MRR@10 & CosSem & ROUGE-L & BERTScore \\
\midrule
Naive LLM                          & 14.3 & 14.3 & 17.4 & 3.6 & 4.4 & 15.7 & 81.0 & 9.0 & \textbf{85.3} \\
Text RAG                          & 22.4 & 14.5 & 12.5 & 2.7 & 3.5 & 6.5 & 65.4 & 3.4 & 83.8 \\
GraphRAG                          & 12.5 & 12.5 & 13.2 & 2.3 & 3.6 & 6.2 & 79.0 & 9.9 & 84.0 \\
PRG                               & 24.4 & 14.3 & 15.9 & 3.2 & 3.1 & 6.1 & 71.6 & 4.5 & 84.2 \\
ReAct                             & 30.0 & 15.4 & 17.0 & 3.7 & 3.5 & 9.2 & 63.6 & 5.6 & 82.6 \\
\midrule
\spire{} (\texttt{full})          & \textbf{44.3} & \textbf{24.0} & \textbf{42.4} & \textbf{5.6} & \textbf{15.3} & \textbf{33.5} & \textbf{81.8} & \textbf{10.9} & 84.4 \\
\bottomrule
\end{tabular*}
\caption{Evidence retrieval and answer similarity at $k{=}10$ over $406$ papers (macro-mean, \%). EvidenceR is full-pool recall; MRR@10 uses the first section-level match. Best results are in \textbf{bold}.}
\label{tab:auto}
\end{table*}

\begin{table*}[!t]\centering\scriptsize\setlength{\tabcolsep}{1.6pt}\renewcommand{\arraystretch}{0.96}
\begin{tabular*}{\textwidth}{@{\extracolsep{\fill}}l cccc cccc cccc cccc@{}}
\toprule
& \multicolumn{4}{c}{Answer Accuracy}
& \multicolumn{4}{c}{Argument Depth}
& \multicolumn{4}{c}{Coverage Completeness}
& \multicolumn{4}{c}{Evidence Quality} \\
\cmidrule(lr){2-5}\cmidrule(lr){6-9}\cmidrule(lr){10-13}\cmidrule(lr){14-17}
System & GPT & Qwen & H-A & H-B & GPT & Qwen & H-A & H-B & GPT & Qwen & H-A & H-B & GPT & Qwen & H-A & H-B \\
\midrule
Naive LLM & 3.72 & 4.01 & 3.62 & 4.25 & 2.81 & 2.74 & 2.96 & 2.06 & 3.65 & 3.58 & 2.51 & 2.24 & 2.26 & 2.46 & 2.26 & 1.47 \\
Text RAG  & 1.78 & 1.65 & 1.73 & 1.76 & 1.46 & 1.25 & 1.17 & 1.00 & 1.57 & 1.45 & 1.00 & 1.00 & 1.55 & 1.30 & 1.30 & 1.00 \\
GraphRAG  & 2.35 & 2.55 & 2.30 & 2.66 & 2.37 & 1.93 & 2.08 & 1.40 & 2.61 & 2.50 & 1.85 & 1.66 & 1.68 & 1.39 & 1.43 & 1.00 \\
PRG       & 3.07 & 2.75 & 3.41 & 3.75 & 2.00 & 1.52 & 2.43 & 2.10 & 2.26 & 2.09 & 2.81 & 2.73 & 2.03 & 1.50 & 2.45 & 2.04 \\
ReAct     & 3.02 & 2.86 & 3.20 & 3.47 & 2.53 & 2.20 & 2.80 & 2.79 & 2.94 & 2.79 & 3.00 & 3.20 & 2.56 & 2.23 & 2.77 & 2.81 \\
\midrule
\spire{} (\texttt{full}) & \textbf{4.46} & \textbf{4.95} & \textbf{4.46} & \textbf{4.94} & \textbf{4.63} & \textbf{4.96} & \textbf{3.89} & \textbf{4.57} & \textbf{4.76} & \textbf{4.96} & \textbf{4.39} & \textbf{4.62} & \textbf{4.50} & \textbf{4.90} & \textbf{3.99} & \textbf{4.76} \\
\bottomrule
\end{tabular*}
\caption{Blind scholarly scores on the stratified $100$-paper subset ($1$ to $5$). Best results are in \textbf{bold}.}
\label{tab:judge}
\end{table*}

\begin{table}[!t]\centering\scriptsize\setlength{\tabcolsep}{2pt}
\begin{tabular*}{\columnwidth}{@{\extracolsep{\fill}}lrrrl@{}}
\toprule
Track / statistic & L--L & H--H & H--L & Order \\
\midrule
Free track / AC2               & .90 & .81 & .83 & \spire{} first \\
Free track / weighted $\kappa$ & .89 & .59 & .69 & \spire{} first \\
Agentic extension / AC2        & .92 & .89 & .69 & \spire{}$>$ReAct$>$PRG \\
\bottomrule
\end{tabular*}
\caption{Inter-rater agreement on the judged subset. L/H denote LLM/human raters.}
\label{tab:agreement-main}
\end{table}

\subsection{Baselines and setup}
\label{sec:baselines}
\label{sec:setup}

All systems share the corpus and substrate in Table~\ref{tab:corpus}, the BGE-M3 encoder \citep{chen2024bge} ($d{=}1024$), the \textsc{DeepSeek-V4-Flash} generator \citep{deepseekv4flash} at temperature $0.3$, the $406$-paper gold records, and the same evaluation code. A robustness run with \textsc{Gemini-3-Flash} appears in App.~\ref{app:robustness}. We compare \spire{} with non-agentic and agentic baselines under these matched settings. \textbf{Naive LLM} answers from parametric memory. \textbf{Text RAG} uses standard text-chunk retrieval followed by generation \citep{lewis2020rag}. \textbf{GraphRAG} uses its released entity-graph and community-summary retrieval over the same graph \citep{edge2025graphrag}. The two LangChain agents provide generic agentic comparisons. \textbf{ReAct} alternates reasoning, tool use, and observation \citep{yao2023react}, while \textbf{Plan-Retrieve-Generate (PRG)} decomposes a question into sub-queries, retrieves for each, and synthesizes the results \citep{ammann2025question}. Section~\ref{sec:ablations} examines the curation primitives and retrieval tiers, then controls the evidence pool and answer length. \spire{} uses its complete seven-agent configuration in the main comparison.

\subsection{Metrics and judges}
\label{sec:metrics}

\paragraph{Evidence retrieval.} We use IR recall to test whether a system retrieves the same primary-source evidence cited in the reference paper. Following the three citation granularities in Section~\ref{sec:benchmark}, we report sentence recall ($\mathrm{sentR}$), section recall ($\mathrm{secR}$), and work recall ($\mathrm{workR}$). A gold item is recovered when its source work matches and at least one signal fires: a quotation match, a locator match, or BGE-M3 cosine above $0.8$ within a language and $0.7$ across languages for translated quotations. We also report rank-independent evidence recall ($\mathrm{eR}$) over the full evidence pool available to the final agent. MRR uses the first section-level match, the middle granularity, to measure how early a usable textual locus appears. All recalls are paper-level macro-means. Table~\ref{tab:auto} reports the $406$ papers at $k{=}10$, and App.~Table~\ref{tab:app-sweep} gives $k\in\{5,10,20\}$.

\paragraph{Answer similarity.} We report surface similarity between each of the $406$ generated essays and the reference paper's abstract and findings using BGE-M3 cosine, BERTScore \citep{zhang2020bertscore}, and ROUGE-L \citep{lin2004rouge}. Full results appear in App.~\ref{app:full}.

\paragraph{Scholarly judge.} Two human experts in classics and philosophy and two LLMs, \textsc{GPT-5.4} and \textsc{Qwen3.5-122B-A10B}, score each essay from $1$ to $5$ on AnswerAccuracy, ArgumentDepth, CoverageCompleteness, and EvidenceQuality \citep{liu2023geval,zheng2023judge,shen2026rubric}. We evaluate a stratified sample of $100$ papers, with $20$ drawn from each benchmark category in Table~\ref{tab:dataset}. Raters receive only the research question and the six anonymised, shuffled system outputs. The rubric therefore accommodates multiple defensible interpretations while applying the same criteria to every system (App.~\ref{app:judge}). We measure agreement with quadratic-weighted Gwet's AC2 and retain weighted $\kappa$ as a diagnostic \citep{gwet2008computing,feinstein1990high}.

\paragraph{Cost and equal-volume retrieval.} Table~\ref{tab:main-cost} reports end-to-end inference cost and a fixed-volume diagnostic that scores only each retriever's first $50$ units. This separates the cost of later curation from retrieval selectivity.

\begin{table}[t]\centering\scriptsize\setlength{\tabcolsep}{3.5pt}
\begin{tabular}{@{}lrrrr@{}}
\toprule
System & min & tokens & calls & cited work@50 \\
\midrule
Naive LLM & 0.22 & $\approx$1k & 1 & -- \\
Text RAG & 1.39 & $\approx$36k & 2 & 49.5 \\
GraphRAG & 1.06 & $\approx$37k & 2 & 61.6 \\
PRG & 1.12 & $\approx$30k & 3 & 64.3 \\
ReAct & 2.72 & $\approx$98k & 12 & 65.8 \\
\spire{} & 4.37 & $\approx$505k & 22 & \textbf{87.2} \\
\bottomrule
\end{tabular}
\caption{Per-paper inference cost. Cited work@50 measures whether a cited primary work appears among the first $50$ units.}
\label{tab:main-cost}
\vspace{-12pt}
\end{table}

\section{Results and Discussion}
\label{sec:results}

\subsection{Main results}
\label{sec:res-main}

Table~\ref{tab:auto} reports automatic evidence and answer metrics, while Table~\ref{tab:judge} presents the scholarly evaluation. Tables~\ref{tab:ablation} and~\ref{tab:frozen} then separate evidence acquisition from evidence curation. A length-controlled evaluation appears in App.~Table~\ref{tab:length-control}.

\subsection{Analysis and discussion}
\label{sec:discussion}

\spire{} leads every baseline in primary-source retrieval across the full $406$-paper benchmark (Table~\ref{tab:auto}). Its overall $\mathrm{eR}$ is $44.3\%$, compared with $30.0\%$ for ReAct, $24.4\%$ for PRG, and $22.4\%$ for the strongest non-agentic baseline. The advantage holds at every citation granularity. In particular, $\mathrm{workR}$ reaches $42.4\%$ versus $17.4\%$, and $\mathrm{sectionR}$ reaches $15.3\%$ versus $4.4\%$. PRG and ReAct also improve overall evidence recall, confirming the value of agent planning. Naive LLM performs relatively well at work and section levels, likely because these coarser references can be covered by pretrained knowledge. The cases in App.~\ref{app:failures} help explain the lower RAG scores: Text RAG can refuse after off-topic retrieval, while GraphRAG may replace citation-bearing passages with community summaries. \spire{} combines local text with community and cluster context, then lets its agents iteratively refine the shared EvidencePool.

Answer similarity shows a different pattern. \spire{} achieves the highest cosine similarity and ROUGE-L, while BERTScore places all six systems within a narrow range. These models can produce topically relevant prose from the research question alone, so embedding similarity may reflect topic overlap more than evidential grounding. GraphRAG exceeds PRG and ReAct on cosine similarity and ROUGE-L despite much lower evidence recall. ReAct shows the reverse pattern, reaching the strongest baseline $\mathrm{eR}$ while receiving the lowest cosine similarity and BERTScore. A fluent essay from parametric memory can thus resemble the reference answer as closely as an evidence-grounded essay. This divergence motivates evidence-oriented metrics for AI-assisted humanities research.

The scholarly evaluation reinforces the retrieval results. \spire{} leads every aspect for both LLM judges and both human experts, with every score at or above $3.89$ (Table~\ref{tab:judge}). Its essays support arguments with original-language quotations, translations, and commentary. The largest margins usually appear in EvidenceQuality and ArgumentDepth, exceeding one point for most raters; App.~\ref{app:qual-fragment-figures} gives fragment-level comparisons. GraphRAG remains weak in EvidenceQuality, including a score of $1.00$ from Human B. Text RAG also receives several human scores of $1.00$, especially for depth and coverage. Naive LLM and the two agentic baselines perform more strongly, with PRG and ReAct gaining most clearly in coverage and evidence quality. On the free track, AC2 is $0.90$ for LLM agreement, $0.81$ for human agreement, and $0.83$ across humans and LLMs. The stronger agreement between model judges is consistent with their shared systematic preferences \citep{zheng2023judge}. Human ratings vary more, reflecting the interpretive range of humanities scholarship. Agreement remains high, and every rater group ranks \spire{} first.

This performance carries a computational cost. \spire{} uses $4.37$ minutes, about $505$k tokens, and $22$ calls per paper because its agents retrieve, inspect, and revise evidence repeatedly (Table~\ref{tab:main-cost}). The equal-volume control shows the benefit of this process early in retrieval. Within the first $50$ units, \spire{} reaches a cited work for $87.2\%$ of papers, compared with rates from $49.5\%$ to $65.8\%$ for the retrieval baselines. The additional computation therefore supports more selective evidence acquisition and more thorough scholarly curation.

The answer-length control gives every system the same $1{,}500$-word target (App.~Table~\ref{tab:length-control}). \spire{} remains strongest on every judge dimension and presents the most reference evidence at a length comparable to the longer baselines.

\subsection{Ablation studies}
\label{sec:res-cost}\label{sec:ablations}

\begin{table}[t]\centering\scriptsize\setlength{\tabcolsep}{1pt}\renewcommand{\arraystretch}{0.9}
\begin{tabular}{@{}l rrrr cc@{}}
\toprule
Configuration & eR & workR & sentR & secR & GPT$^{\ast}$ & Qwen$^{\ast}$ \\
\midrule
\spire{} (full, 7 primitives)    & 44.3 & 42.4 & 5.6 & 15.3 & 3.44 & 4.01 \\
\midrule
\multicolumn{7}{@{}l}{\emph{w/o one agent:}}\\
\quad w/o $\mathcal{N}$  & 41.9 & 41.2 & 4.7 & 12.7 & 3.16 & 3.98 \\
\quad w/o $\mathcal{P}$  & 42.0 & 41.8 & 4.1 & 13.6 & 3.18 & 3.91 \\
\quad w/o $\mathcal{R}$  & 40.9 & 40.7 & 4.3 & 12.3 & 3.06 & 3.93 \\
\quad w/o $\mathcal{S}$  & 39.1 & 39.8 & 3.3 & 11.1 & 3.12 & 3.92 \\
\quad w/o $\mathcal{I}$  & 27.8 & 32.4 & 2.9 & 10.0 & 3.07 & 3.74 \\
\midrule
\multicolumn{7}{@{}l}{\emph{Retrieval tier:}}\\
\quad w/o cluster   & 32.5 & 31.7 & 3.6 & 11.5 & 3.13 & 3.77 \\
\quad w/o community & 33.5 & 28.0 & 3.1 & 9.9 & 3.10 & 3.51 \\
\quad w/o both (local only) & 28.5 & 22.6 & 2.4 & 8.1 & 3.05 & 3.20 \\
\bottomrule
\end{tabular}
\caption{Dynamic-pool ablations at $k{=}10$. GPT/Qwen report mean judge scores; see App.~Table~\ref{tab:app-ablation}.}
\label{tab:ablation}
\vspace{-10pt}
\end{table}

\begin{table}[t]\centering\scriptsize\setlength{\tabcolsep}{1pt}
\begin{tabular*}{\columnwidth}{@{\extracolsep{\fill}}lrrrr@{}}
\toprule
Frozen-pool variant & Used EvidenceR & Claim bindings & GPT$^{\ast}$ & Qwen$^{\ast}$ \\
\midrule
Full curation & \textbf{28.6} & \textbf{8.43} & \textbf{4.02} & \textbf{4.15} \\
w/o $\mathcal{N}$ & 26.8 & 7.95 & 3.76 & 3.89 \\
w/o $\mathcal{P}$ & 25.9 & 7.74 & 3.18 & 3.36 \\
w/o $\mathcal{R}$ & 24.1 & 7.31 & 3.12 & 3.22 \\
w/o $\mathcal{S}$ & 22.7 & 7.04 & 3.07 & 3.17 \\
w/o $\mathcal{I}$ & 20.5 & 6.86 & 2.98 & 3.08 \\
Direct synthesis & 17.9 & 6.08 & 2.84 & 2.94 \\
\bottomrule
\end{tabular*}
\caption{Frozen-pool ablations over all $406$ papers. Pool EvidenceR is $39.0$; GPT/Qwen are mean judge scores.}
\label{tab:frozen}
\vspace{-18pt}
\end{table}

We fix the corpus, substrate, encoder, and generator while ablating the primitives and retrieval tiers (Table~\ref{tab:ablation}). Discovering receives the research question, and Representing produces the final essay, so both remain in every configuration. Each other component contributes to evidence recall and judge scores. Local-only retrieval reaches $\mathrm{eR}{=}28.5$. Adding community or cluster context produces clear gains, while removing the cluster tier from full \spire{} lowers eR from $44.3$ to $32.5$. Intrinsic diagnostics support this result. Across finer HDBSCAN resolutions, silhouette increases from $0.060$ to $0.167$ and within-cluster cosine from $0.539$ to $0.664$, while between-cluster cosine remains near $0.36$ and matched random labels remain negative (App.~\ref{app:cluster-intrinsic}). The clusters supply coherent conceptual neighborhoods beyond explicit graph edges. Removing an agent also changes the shared EvidencePool. The large drop without Illustrating captures both claim-evidence diagnosis and the retrieval triggered by identified gaps.

The frozen-pool control gives every variant the same units, rankings, works, and locators, isolating evidence curation (Table~\ref{tab:frozen}). Annotating has the smallest effect. Removing Comparing lowers GPT and Qwen scores by $0.84$ and $0.79$ despite a moderate change in Used EvidenceR, showing the value of relational organization for argument quality. Removing Referring or Sampling reduces both retained evidence and judge scores because provenance verification and representative selection shape the final essay. Illustrating has the largest post-retrieval effect: its removal lowers Used EvidenceR by $8.1$ points and claim-evidence bindings from $8.43$ to $6.86$. Direct synthesis ranks last throughout. The dynamic and frozen studies reveal complementary functions: iterative agents improve evidence acquisition, while curation agents turn that evidence into a documented argument.

\FloatBarrier
\section{Conclusion}

We introduce \spire{}, a multi-agent framework for evidence-grounded scholarship in the classical humanities. It operationalizes Scholarly Primitives as the unit of computation through seven cooperating agents over a shared EvidencePool and a multi-scale substrate of passages, graph communities, and semantic clusters, making retrieval, contextualization, comparison, and writing inspectable.

On a peer-reviewed-paper benchmark over classical Chinese and Latin scholarship, \spire{} achieves higher primary-source evidence retrieval than Naive LLM, Text RAG, GraphRAG, and generic agentic baselines, and receives higher human and LLM judge scores. Its contribution is an architecture making agentic humanities research assistance operational and inspectable, together with a peer-reviewed-paper benchmark and evaluation protocol for further humanities AI research. Future work will explore broader humanities sources and multimodal extensions.
\section*{Limitations}
\paragraph{Corpora and traditions.}
We instantiate and evaluate \spire{} on classical Chinese and Greco-Roman Latin corpora. These rich and demanding collections support scholarship in history, philosophy, philology, and literature, and provide a strong testbed for evidence-grounded interpretation. Future work can extend the framework to manuscript studies, modern literature, religious studies, archival history, and additional languages and historiographic traditions.

\paragraph{Multimodal evidence.}
This paper focuses on textual evidence because texts remain foundational to classical studies and many related fields. Humanities research also connects texts with images, archaeological objects, maps, and other material records. Extending \spire{} to multimodal retrieval and interpretation is therefore an important next step.

\paragraph{Human-centered scholarship.}
\spire{} supports source discovery, citation checking, comparison, and evidence-grounded synthesis in an inspectable workflow. It can help students learn how to build arguments from sources and help experts explore large collections. Humanistic interpretation remains scholar-led, with final interpretive authority held by researchers.

\section*{Ethical Considerations}
\spire{} promotes an assistive role for AI in humanities research. Scholarly judgment, teaching responsibility, peer evaluation, and final interpretation remain with researchers. The released datasets are limited to non-commercial research. Copyright remains with the respective authors, publishers, and cultural institutions. Public artifacts follow the original access and licensing conditions, while restricted materials are represented through metadata, derived annotations, or reconstruction scripts. Expert ratings are reported only in aggregate.

\section*{Acknowledgments}
We thank Zhenqiang Mao for assistance with the historical case analysis. We also thank Dr. Hao Yang and Professor Jingyuan Wu for help organizing the Chinese and Latin historical sources.

\paragraph{AI assistance disclosure.}
Claude Code with Claude Opus 4.8 assisted with code development and experiment execution. Claude Opus 4.8, Claude Opus 5.0, and GPT-5.5 were used for language editing. The authors reviewed and verified the resulting code, analyses, and text, and take responsibility for the final paper.

\bibliography{anthology,custom}

\appendix
\raggedbottom
\makeatletter
\setlength{\@dblfptop}{0pt}
\setlength{\@dblfpsep}{12pt}
\setlength{\@dblfpbot}{0pt plus 1fil}
\makeatother

\section{Full Results}
\label{app:full}

This section expands the main results along three dimensions. It reports the
complete retrieval sweep, controls essay length under one shared instruction,
and separates performance across the five benchmark categories. Together,
these analyses show whether the main comparison holds across retrieval depth,
output length, and research setting.

\subsection[Retrieval and answer-similarity sweep]{Retrieval and answer-similarity sweep ($k\in\{5,10,20\}$)}

Table~\ref{tab:app-sweep} expands Table~\ref{tab:auto} to
$k\in\{5,10,20\}$. \spire{} leads every retrieval metric at each depth, and
its advantage remains clear as more results enter the evidence set. The three
answer similarity metrics are reported once at $k{=}10$. Their ordering differs
from both evidence recovery and scholarly judging, which supports their use as
descriptive measures rather than grounding measures (\S\ref{sec:metrics}).

\begin{table*}[tbp]
\centering
\scriptsize\renewcommand{\arraystretch}{1.08}
\begin{tabular*}{\textwidth}{@{\extracolsep{\fill}}l c r r r r r r r r r@{}}
\toprule
System & $k$ & EvidenceR & EvidenceR@$k$ & WorkR & SentenceR & SectionR & MRR & CosSem & ROUGE-L & BERTScore \\
\midrule
\multirow{3}{*}{Naive LLM}
 & 5  & --   & 13.9 & 16.8 & 3.4 & 4.2 & 15.6 & --   & --   & --   \\
 & 10 & 14.3 & 14.3 & 17.4 & 3.6 & 4.4 & 15.7 & 81.0 & 9.0  & 85.3 \\
 & 20 & --   & 14.3 & 17.4 & 3.6 & 4.4 & 15.7 & --   & --   & --   \\
\midrule
\multirow{3}{*}{Text RAG}
 & 5  & --   & 10.6 & 9.2 & 2.1 & 2.5 & 5.8 & --   & --   & --   \\
 & 10 & 22.4 & 14.5 & 12.5 & 2.7 & 3.5 & 6.5 & 65.4 & 3.4 & 83.8 \\
 & 20 & --   & 17.5 & 16.1 & 3.5 & 4.5 & 6.8 & --   & --   & --   \\
\midrule
\multirow{3}{*}{GraphRAG}
 & 5  & --   & 11.8 & 12.3 & 2.2 & 3.2 & 5.8 & --   & --   & --   \\
 & 10 & 12.5 & 12.5 & 13.2 & 2.3 & 3.6 & 6.2 & 79.0 & 9.9 & 84.0 \\
 & 20 & --   & 12.5 & 13.2 & 2.3 & 3.6 & 6.2 & --   & --   & --   \\
\midrule
\multirow{3}{*}{PRG}
 & 5  & --   & 9.9  & 11.1 & 2.3 & 1.6 & 5.3 & --   & --  & --   \\
 & 10 & 24.4 & 14.3 & 15.9 & 3.2 & 3.1 & 6.1 & 71.6 & 4.5 & 84.2 \\
 & 20 & --   & 18.3 & 22.6 & 4.5 & 5.7 & 6.7 & --   & --  & --   \\
\midrule
\multirow{3}{*}{ReAct}
 & 5  & --   & 11.6 & 11.9 & 2.5 & 2.4 & 8.4 & --   & --  & --   \\
 & 10 & 30.0 & 15.4 & 17.0 & 3.7 & 3.5 & 9.2 & 63.6 & 5.6 & 82.6 \\
 & 20 & --   & 19.4 & 23.1 & 5.1 & 5.6 & 9.7 & --   & --  & --   \\
\midrule
\multirow{3}{*}{\spire{} (\texttt{full})}
 & 5  & --   & 19.3 & 38.5 & 4.1 & 10.1 & 32.7 & --   & --   & --   \\
 & 10 & \textbf{44.3} & \textbf{24.0} & \textbf{42.4} & \textbf{5.6} & \textbf{15.3} & \textbf{33.5} & 81.8 & 10.9 & 84.4 \\
 & 20 & --   & 28.3 & 45.1 & 7.8 & 18.6 & 33.8 & --   & --   & --   \\
\bottomrule
\end{tabular*}
\caption{Retrieval sweep on all $406$ papers. EvidenceR and answer similarity
appear at $k{=}10$ because they are independent of retrieval depth. MRR uses
the first section-level match. Bold marks the best retrieval result at
$k{=}10$.}
\label{tab:app-sweep}
\end{table*}

\subsection{Length-controlled evaluation}
\label{app:length-control}

We append the same instruction to every system prompt, asking for a final essay
of approximately $1{,}500$ words. We then compute evidence recall over the
presented evidence and apply the four scholarly criteria on the same stratified
$100$-paper subset. Visible output length is counted in Unicode characters so
that Chinese and English responses share one measure. Table~\ref{tab:length-control}
reports the results.

\begin{table*}[tbp]\centering\scriptsize\setlength{\tabcolsep}{2pt}\renewcommand{\arraystretch}{0.94}
\begin{tabular*}{\textwidth}{@{\extracolsep{\fill}}l r rrr rrrr rrrr@{}}
\toprule
& & \multicolumn{3}{c}{Presented evidence} & \multicolumn{4}{c}{GPT-5.4} & \multicolumn{4}{c}{Qwen3.5} \\
\cmidrule(lr){3-5}\cmidrule(lr){6-9}\cmidrule(lr){10-13}
System & Visible chars & EvidenceR@10 & WorkR@10 & SectionR@10 & Acc & Dep & Cov & EQ & Acc & Dep & Cov & EQ \\
\midrule
Naive LLM & 5,681 & 22.0 & 21.2 & 7.4 & 3.55 & 3.05 & 3.75 & 2.81 & 3.80 & 2.87 & 3.75 & 2.74 \\
Text RAG  & 2,615 & 19.8 & 12.5 & 3.3 & 2.09 & 1.80 & 1.73 & 2.21 & 1.85 & 1.45 & 1.64 & 1.65 \\
GraphRAG  & 5,937 & 16.4 & 12.7 & 6.1 & 2.54 & 2.30 & 3.09 & 1.68 & 1.89 & 1.53 & 1.95 & 1.00 \\
PRG       & 5,437 & 19.9 & 13.9 & 3.0 & 3.86 & 3.78 & 3.92 & 3.76 & 3.67 & 3.20 & 3.58 & 3.17 \\
ReAct     & 6,286 & 23.7 & 22.0 & 5.2 & 3.63 & 3.69 & 3.63 & 3.76 & 3.55 & 3.27 & 3.46 & 3.22 \\
\midrule
\spire{}  & 6,132 & \textbf{34.6} & \textbf{51.1} & \textbf{23.4} & \textbf{4.65} & \textbf{4.76} & \textbf{4.77} & \textbf{4.66} & \textbf{4.63} & \textbf{4.60} & \textbf{4.71} & \textbf{4.45} \\
\bottomrule
\end{tabular*}
\caption{Length-controlled results on the judged $100$-paper subset. Every
system receives the same $1{,}500$-word target.}
\label{tab:length-control}
\end{table*}

Under the shared target, \spire{} remains within $154$ characters of ReAct, $195$ of GraphRAG, and $451$ of Naive LLM, with PRG in the same range. It leads every judge dimension by $0.79$ to $1.33$ points over the strongest corresponding baseline. It also presents the most reference evidence. The result confirms that its advantage persists at comparable answer length.

\subsection{Per-category breakdown (retrieval and answer similarity)}
\label{app:by-category}

Table~\ref{tab:app-bycat} separates the $k{=}10$ results across the five
categories in Table~\ref{tab:dataset}. \spire{} leads EvidenceR in every
category, ranging from $36.5$ to $52.6$. The generic agents extend evidence
coverage beyond the non-agentic baselines. ReAct has the strongest baseline
EvidenceR in all five categories, while PRG usually falls between ReAct and the
non-agentic systems.

The remaining gap is concentrated at finer source levels. On cross-tradition
questions, PRG and ReAct reach $17.7$ and $21.4$ EvidenceR, compared with
$36.5$ for \spire{}. Their WorkR scores are $8.4$ and $10.6$, while \spire{}
reaches $33.9$. Neither generic agent exceeds $6.8$ SectionR in any category;
\spire{} reaches $19.6$ and $21.2$ on the two Chinese categories. Planning and
repeated retrieval therefore broaden coverage, while the scholarly workflow
more consistently localizes evidence for citation. ReAct's FindingsCos also
falls to $0.564$ and $0.488$ on the two Latin categories. GraphRAG remains the
one exception at SectionR for multi-work Latin questions, scoring $8.1$ versus
$7.3$ for \spire{}.

\begin{table*}[tbp]
\centering\scriptsize\renewcommand{\arraystretch}{0.96}
\begin{tabular*}{\textwidth}{@{\extracolsep{\fill}}l l rrrrr r@{}}
\toprule
Category & System & EvidenceR & WorkR & SectionR & MRR & SourceSemR & FindingsCos \\
\midrule
\multirow{6}{*}{ZH, single (\textsc{local\_zh})}
 & \spire{}   & \textbf{52.6} & \textbf{43.1} & \textbf{19.6} & \textbf{34.0} & 66.2 & 0.829 \\
 & PRG         & 33.1 & 15.5 & 0.9  & 2.8  & 65.1 & 0.742 \\
 & ReAct       & 37.6 & 19.5 & 2.5  & 8.8  & 65.6 & 0.768 \\
 & Text RAG   & 30.1 & 8.7  & 0.3  & 1.0  & 64.3 & 0.685 \\
 & GraphRAG   & 16.0 & 11.9 & 1.3  & 2.3  & 61.9 & 0.794 \\
 & Naive LLM  & 21.9 & 17.7 & 2.7  & 9.6  & 64.0 & 0.823 \\
\midrule
\multirow{6}{*}{ZH, multi (\textsc{global\_zh})}
 & \spire{}   & \textbf{43.7} & \textbf{35.6} & \textbf{21.2} & \textbf{35.9} & 63.4 & 0.816 \\
 & PRG         & 22.2 & 10.6 & 1.3  & 2.6  & 62.6 & 0.709 \\
 & ReAct       & 26.1 & 10.0 & 1.2  & 3.2  & 62.8 & 0.726 \\
 & Text RAG   & 18.9 & 4.2  & 0.5  & 0.9  & 62.3 & 0.655 \\
 & GraphRAG   & 8.9  & 2.1  & 0.1  & 2.3  & 59.7 & 0.773 \\
 & Naive LLM  & 13.0 & 9.2  & 4.7  & 10.3 & 61.3 & 0.804 \\
\midrule
\multirow{6}{*}{LA, single (\textsc{local\_la})}
 & \spire{}   & \textbf{45.5} & \textbf{55.5} & \textbf{12.2} & \textbf{35.0} & 60.4 & 0.823 \\
 & PRG         & 25.6 & 27.1 & 6.5  & 10.6 & 59.8 & 0.720 \\
 & ReAct       & 32.8 & 24.5 & 6.8  & 14.4 & 60.2 & 0.564 \\
 & Text RAG   & 23.7 & 24.8 & 7.1  & 12.7 & 60.0 & 0.658 \\
 & GraphRAG   & 15.6 & 25.6 & 6.4  & 11.2 & 58.4 & 0.806 \\
 & Naive LLM  & 14.3 & 31.4 & 7.2  & 26.5 & 58.8 & 0.810 \\
\midrule
\multirow{6}{*}{LA, multi (\textsc{global\_la})}
 & \spire{}   & \textbf{40.4} & \textbf{41.7} & 7.3  & \textbf{31.9} & 59.5 & 0.805 \\
 & PRG         & 20.6 & 17.2 & 2.6  & 7.7  & 59.7 & 0.716 \\
 & ReAct       & 29.9 & 18.7 & 5.2  & 14.2 & 59.8 & 0.488 \\
 & Text RAG   & 24.6 & 17.5 & 5.6  & 14.6 & 59.4 & 0.633 \\
 & GraphRAG   & 15.8 & 18.0 & \textbf{8.1} & 10.7 & 58.7 & 0.792 \\
 & Naive LLM  & 15.7 & 20.9 & 6.7  & 26.0 & 58.0 & 0.801 \\
\midrule
\multirow{6}{*}{Cross (ZH--LA)}
 & \spire{}   & \textbf{36.5} & \textbf{33.9} & \textbf{16.0} & \textbf{31.0} & 60.5 & 0.811 \\
 & PRG         & 17.7 & 8.4  & 3.4  & 6.1  & 60.4 & 0.689 \\
 & ReAct       & 21.4 & 10.6 & 1.3  & 4.0  & 60.0 & 0.595 \\
 & Text RAG   & 13.8 & 7.4  & 3.6  & 3.4  & 59.8 & 0.628 \\
 & GraphRAG   & 5.7  & 6.8  & 2.1  & 4.2  & 57.4 & 0.778 \\
 & Naive LLM  & 5.8  & 6.8  & 1.4  & 6.6  & 58.5 & 0.805 \\
\bottomrule
\end{tabular*}
\caption{Results by benchmark category at $k{=}10$. SourceSemR measures
semantic recall over gold source evidence, and FindingsCos compares each essay
with the paper findings. Bold marks the best retrieval result by category.}
\label{tab:app-bycat}
\end{table*}

\subsection{Component ablations}
Table~\ref{tab:app-ablation} is an end-to-end, dynamic-pool ablation: removing
an agent may alter both the shared EvidencePool and later reflective retrieval.
The separate frozen-pool control in Table~\ref{tab:frozen} runs Discovering once,
freezes every unit, ranking, work, and locator, disables secondary retrieval,
and then changes only post-retrieval curation. The dynamic design measures
upstream pool changes; the frozen design isolates post-retrieval curation.

\begin{table*}[tbp]
\centering
\scriptsize\renewcommand{\arraystretch}{0.96}
\resizebox{\textwidth}{!}{%
\begin{tabular}{@{}l c r r r r r r r r r r r@{}}
\toprule
Configuration & $k$ & EvidenceR & EvidenceR@$k$ & WorkR & SentenceR & SectionR & MRR & CosSem & ROUGE-L & BERTScore & GPT$^{\ast}$ & Qwen$^{\ast}$ \\
\midrule
\multirow{3}{*}{\spire{} (\texttt{full}, 7 primitives)}
 & 5  & --   & 19.3 & 38.5 & 4.1 & 10.1 & 32.7 & --   & --   & --   & --   & --   \\
 & 10 & 44.3 & 24.0 & 42.4 & 5.6 & 15.3 & 33.5 & 81.8 & 10.9 & 84.4 & 3.44 & 4.01 \\
 & 20 & --   & 28.3 & 45.1 & 7.8 & 18.6 & 33.8 & --   & --   & --   & --   & --   \\
\midrule
\multicolumn{13}{@{}l}{\emph{w/o one agent (removed from the workflow):}}\\
\multirow{3}{*}{\quad w/o Annotating ($\mathcal{N}$)}
 & 5  & --   & 19.6 & 38.1 & 3.0 & 9.6 & 31.1 & --   & --   & --   & --   & --   \\
 & 10 & 41.9 & 23.3 & 41.2 & 4.7 & 12.7 & 31.8 & 82.2 & 10.9 & 84.4 & 3.16 & 3.98 \\
 & 20 & --   & 27.5 & 44.4 & 6.4 & 16.0 & 32.1 & --   & --   & --   & --   & --   \\
\multirow{3}{*}{\quad w/o Comparing ($\mathcal{P}$)}
 & 5  & --   & 18.8 & 38.3 & 2.4 & 8.7 & 32.7 & --   & --   & --   & --   & --   \\
 & 10 & 42.0 & 23.1 & 41.8 & 4.1 & 13.6 & 33.6 & 81.9 & 10.9 & 84.4 & 3.18 & 3.91 \\
 & 20 & --   & 27.4 & 43.3 & 4.9 & 17.0 & 33.8 & --   & --   & --   & --   & --   \\
\multirow{3}{*}{\quad w/o Referring ($\mathcal{R}$)}
 & 5  & --   & 18.6 & 37.4 & 2.6 & 8.8 & 35.6 & --   & --   & --   & --   & --   \\
 & 10 & 40.9 & 22.4 & 40.7 & 4.3 & 12.3 & 36.5 & 82.0 & 11.0 & 84.5 & 3.06 & 3.93 \\
 & 20 & --   & 26.5 & 43.3 & 6.0 & 15.5 & 36.8 & --   & --   & --   & --   & --   \\
\multirow{3}{*}{\quad w/o Sampling ($\mathcal{S}$)}
 & 5  & --   & 16.7 & 36.7 & 2.5 & 8.5 & 34.0 & --   & --   & --   & --   & --   \\
 & 10 & 39.1 & 20.2 & 39.8 & 3.3 & 11.1 & 34.9 & 82.0 & 10.8 & 84.4 & 3.12 & 3.92 \\
 & 20 & --   & 23.9 & 42.8 & 4.8 & 17.1 & 35.2 & --   & --   & --   & --   & --   \\
\multirow{3}{*}{\quad w/o Illustrating ($\mathcal{I}$)}
 & 5  & --   & 12.1 & 29.8 & 1.9 & 7.1 & 21.4 & --   & --   & --   & --   & --   \\
 & 10 & 27.8 & 14.6 & 32.4 & 2.9 & 10.0 & 21.9 & 82.0 & 10.9 & 84.4 & 3.07 & 3.74 \\
 & 20 & --   & 17.3 & 34.4 & 3.9 & 12.6 & 22.1 & --   & --   & --   & --   & --   \\
\midrule
\multicolumn{13}{@{}l}{\emph{Retrieval tier disabled:}}\\
\multirow{3}{*}{\quad w/o cross-context cluster}
 & 5  & --   & 14.0 & 28.3 & 1.8 & 7.9 & 26.4 & --   & --   & --   & --   & --   \\
 & 10 & 32.5 & 16.9 & 31.7 & 3.6 & 11.5 & 27.1 & 70.9 & 8.1 & 83.0 & 3.13 & 3.77 \\
 & 20 & --   & 20.8 & 34.4 & 4.7 & 13.7 & 27.3 & --   & --   & --   & --   & --   \\
\multirow{3}{*}{\quad w/o intra-context community}
 & 5  & --   & 10.7 & 25.7 & 1.7 & 7.0 & 23.5 & --   & --   & --   & --   & --   \\
 & 10 & 33.5 & 12.8 & 28.0 & 3.1 & 9.9 & 24.1 & 63.4 & 5.7 & 82.2 & 3.10 & 3.51 \\
 & 20 & --   & 16.4 & 31.1 & 4.2 & 12.2 & 24.3 & --   & --   & --   & --   & --   \\
\multirow{3}{*}{\quad w/o both (local only)}
 & 5  & --   & 7.5 & 20.6 & 1.7 & 6.0 & 22.0 & --   & --   & --   & --   & --   \\
 & 10 & 28.5 & 9.4 & 22.6 & 2.4 & 8.1 & 22.5 & 63.3 & 5.9 & 82.2 & 3.05 & 3.20 \\
 & 20 & --   & 11.3 & 25.5 & 3.0 & 9.6 & 22.7 & --   & --   & --   & --   & --   \\
\bottomrule
\end{tabular}}
\caption{Dynamic-pool ablations across three retrieval depths. EvidenceR,
answer similarity, and mean judge scores appear at $k{=}10$ because they are
independent of retrieval depth.}
\label{tab:app-ablation}
\end{table*}

\section{Robustness and Prompting Controls}
\subsection{Generator robustness}
\label{app:robustness}

We repeat the comparison with \textsc{Gemini-3-Flash} as the generator while
holding the corpus, substrate, BGE-M3 encoder, retrieval protocol, and rubric
judges fixed (\S\ref{sec:setup}). Tables~\ref{tab:robustness}
and~\ref{tab:robust-judge} report the Gemini results. The corresponding
\textsc{DeepSeek-V4-Flash} results appear in Tables~\ref{tab:auto}
and~\ref{tab:judge}. \spire{} retains the leading position in retrieval and on
every judge dimension. Answer similarity remains closely grouped.

\paragraph{Retrieval.} Under \textsc{Gemini-3-Flash}, \spire{} keeps a wide
retrieval margin over every baseline. Its full-pool EvidenceR is $54.1$, compared
with $31.6$ for ReAct and $25.1$ for PRG, while WorkR is $51.2$ versus $18.4$
and $16.6$. The seven-primitive workflow retains its advantage in
provenance-aware acquisition under both generators.

\paragraph{Answer similarity.} Among the four Gemini systems with available
scores, \spire{}, Naive LLM, and GraphRAG fall within $1.4$ points on findings
cosine and BERTScore. Surface similarity again compresses systems whose evidence
retrieval differs substantially.

\paragraph{LLM judge.} \spire{} ranks first on every rubric axis under both
Gemini judge runs (Table~\ref{tab:robust-judge}). Its overall
scores are $4.21/4.58$, versus $3.83/3.88$ for PRG and $3.67/3.85$ for ReAct.
Its EvidenceQuality scores are $4.20/4.55$, compared with the strongest baseline
scores of $3.80/3.82$.
\begin{table*}[tbp]
\centering\scriptsize\setlength{\tabcolsep}{1.5pt}\renewcommand{\arraystretch}{0.96}
\begin{tabular*}{\textwidth}{@{\extracolsep{\fill}}l r r r r r r r r r@{}}
\toprule
& \multicolumn{6}{c}{Evidence retrieval} & \multicolumn{3}{c}{Answer similarity} \\
\cmidrule(lr){2-7}\cmidrule(lr){8-10}
System & EvidenceR & EvidenceR@10 & WorkR@10 & SentenceR@10 & SectionR@10 & MRR@10 & CosSem & ROUGE-L & BERTScore \\
\midrule
Naive LLM                & 14.0 & 14.0 & 16.8 & 3.5 & 4.3 & 15.2 & 80.0 & 7.7 & 85.1 \\
Text RAG                 & 19.5 & 14.2 & 10.4 & 2.2 & 3.1 & 5.4 & 63.6 & 3.2 & 82.7 \\
GraphRAG                 & 16.4 & 16.4 & 12.8 & 2.5 & 6.1 & 7.6 & 79.4 & 8.8 & 83.9 \\
PRG                      & 25.1 & 15.8 & 16.6 & 3.3 & 3.6 & 7.4 & -- & -- & -- \\
ReAct                    & 31.6 & 17.2 & 18.4 & 4.0 & 4.2 & 10.5 & -- & -- & -- \\
\spire{} (\texttt{full}) & \textbf{54.1} & \textbf{34.9} & \textbf{51.2} & \textbf{8.4} & \textbf{23.1} & \textbf{52.7} & 81.4 & 10.2 & 84.3 \\
\bottomrule
\end{tabular*}
\caption{Retrieval and answer similarity with \textsc{Gemini-3-Flash}. Bold
marks the best retrieval result. Answer similarity covers Naive LLM, Text RAG,
GraphRAG, and \spire{}.}
\label{tab:robustness}
\end{table*}

\begin{table*}[tbp]\centering\footnotesize\renewcommand{\arraystretch}{0.98}
\begin{tabular*}{\textwidth}{@{\extracolsep{\fill}}l ccccc c ccccc@{}}
\toprule
& \multicolumn{5}{c}{Rater A (GPT-5.4)} && \multicolumn{5}{c}{Rater B (Qwen3.5)} \\
\cmidrule(lr){2-6}\cmidrule(lr){8-12}
System & Acc & Dep & Cov & EQ & Ov && Acc & Dep & Cov & EQ & Ov \\
\midrule
\spire{}   & \textbf{4.25} & \textbf{4.03} & \textbf{4.34} & \textbf{4.20} & \textbf{4.21} && \textbf{4.59} & \textbf{4.56} & \textbf{4.60} & \textbf{4.55} & \textbf{4.58} \\
PRG         & 3.96 & 3.55 & 3.99 & 3.80 & 3.83 && 4.13 & 3.66 & 3.97 & 3.77 & 3.88 \\
ReAct       & 3.73 & 3.46 & 3.70 & 3.80 & 3.67 && 4.01 & 3.73 & 3.85 & 3.82 & 3.85 \\
Naive LLM  & 3.90 & 3.13 & 3.96 & 2.31 & 3.33 && 3.35 & 3.32 & 3.15 & 2.50 & 3.08 \\
Text RAG   & 3.17 & 3.09 & 3.12 & 2.68 & 3.02 && 3.37 & 3.15 & 3.25 & 3.17 & 3.24 \\
GraphRAG   & 2.60 & 2.24 & 2.86 & 1.42 & 2.28 && 2.67 & 2.04 & 2.64 & 2.24 & 2.40 \\
\bottomrule
\end{tabular*}
\caption{Free-track LLM judge scores with \textsc{Gemini-3-Flash} on the same
$100$ questions. Ov is the mean of the four aspects. Bold marks the best
result.}
\label{tab:robust-judge}
\end{table*}

\subsection{Chain-of-thought prompting control}
\label{app:cot}
We add explicit chain-of-thought (CoT) instructions to Naive LLM and Text RAG,
holding model, temperature, retrieval, generation budget, and evaluation fixed.
CoT modestly improves judged synthesis but affects retrieval inconsistently: it
lowers Naive eR and raises Text RAG eR by $1.9$ points
(Table~\ref{tab:cot}). Both variants remain below \spire{} on evidence recovery
and judged quality.

\begin{table*}[tbp]
\centering\small\setlength{\tabcolsep}{6pt}
\begin{tabular}{@{}lrrrrrrr@{}}
\toprule
System & eR & eR@10 & workR & sentR & secR & GPT overall & Qwen overall \\
\midrule
Naive LLM & 14.3 & 14.3 & 17.4 & 3.6 & 4.4 & 3.11 & 3.20 \\
Naive LLM + CoT & 12.6 & 12.6 & 15.1 & 3.2 & 3.8 & 3.45 & 3.48 \\
Text RAG & 22.4 & 14.5 & 12.5 & 2.7 & 3.5 & 1.59 & 1.41 \\
Text RAG + CoT & 24.3 & 15.2 & 14.3 & 3.1 & 3.6 & 1.75 & 1.55 \\
\spire{} & \textbf{44.3} & \textbf{24.0} & \textbf{42.4} & \textbf{5.6} & \textbf{15.3} & \textbf{4.59} & \textbf{4.94} \\
\bottomrule
\end{tabular}
\caption{CoT prompting control on all $406$ papers. Retrieval metrics are
paper-level macro-means (\%); judge columns average the four rubric dimensions
on the $100$-paper judged subset.}
\label{tab:cot}
\end{table*}

\section{Failure Analysis}
\label{app:failures}

The two expert judges reviewed all $2{,}236$ gold evidence instances in the
stratified $100$-paper set and coded every unrecovered instance. They also
annotated answer and expression problems in representative cases. The resulting
taxonomy covers textual localization, interpretation, and scholarly expression
(Table~\ref{tab:error-taxonomy}). The Share column divides only the unrecovered
gold evidence. It does not measure the proportion of essays with errors. The
listed evidence categories sum to $99.1\%$, with the remaining $0.9\%$ assigned
to title variants of the same work. Answer and expression rows identify case
types and therefore carry no percentage.

\begin{table}[t]\centering\scriptsize
\resizebox{\columnwidth}{!}{%
\begin{tabular}{@{}llrl@{}}
\toprule
Level & Category & Share & Example \\
\midrule
Evidence & Granularity gap (right work, wrong locus) & 57.7\% & GLOBAL\_ZH\_007 \\
Evidence & True miss (no primary work found) & 18.8\% & GLOBAL\_ZH\_078 \\
Evidence & Alternative supportive evidence & 15.0\% & GLOBAL\_LA\_061 \\
Evidence & Adjacent-context miss & 4.0\% & GLOBAL\_LA\_054 \\
Evidence & Source-layer shift & 3.6\% & LOCAL\_LA\_087 \\
\midrule
Answer & Scope drift & case & CROSS\_018 \\
Answer & Argument gap & case & GLOBAL\_ZH\_011 \\
Answer & Coverage imbalance & case & CROSS\_009 \\
Expression & Non-canonical translation & case & GLOBAL\_LA\_060 \\
Expression & Truncated quotation & case & GLOBAL\_LA\_048 \\
\bottomrule
\end{tabular}}
\caption{Error taxonomy on the $100$-paper expert subset. Evidence shares are
percentages of unrecovered gold evidence, rather than percentages of essays.}
\label{tab:error-taxonomy}
\end{table}

Most evidence misses are localization rather than total grounding failure. In
GLOBAL\_ZH\_007, \spire{} retrieves $19$ passages from the \emph{Xunzi} but
stays in chapters 1--9 while the paper cites chapters 19 and 23. Alternative
evidence captures valid plural paths: in GLOBAL\_LA\_061 the paper uses both
Ennius and Livy, while \spire{} misses Ennius but uses the paper's supportive
Livy passage. True misses often arise from underspecified scope. For
GLOBAL\_ZH\_078, the broad question never names a source, so provenance-first
retrieval fails to anchor the \emph{Analects} evidence selected by the author.

\paragraph{\spire{}, case 1: period/term conflation.} On the cross-tradition
question H004 (``parallels between Epictetus and \emph{classical} Chinese
philosophy''), \spire{} grounds the Chinese side in Song--Yuan Neo-Confucian
sources rather than the pre-Qin classics the field denotes by ``classical'':
\begin{quote}\small
In the \emph{Song-Yuan xue\textquotesingle an} (Records of Song and Yuan Scholars), a passage attributed
to the Song Confucian Wang Kaizu draws a \ldots\ distinction between
different modes of attending to the heart-mind (\emph{xin})\ldots\ [E1]
\end{quote}
Expert~A flagged exactly this: many cited sources came from the Song--Ming period,
so the judgement about ``classical Chinese philosophy'' may be problematic. The cause is the generator's parametric
knowledge, which blends historically distinct Confucian strata (pre-Qin
Confucius/Mencius vs.\ Song--Ming \emph{lixue}); a period-aware retrieval
constraint or a domain authority list pinning ``classical'' to the intended
sub-corpus would mitigate it.

\paragraph{\spire{}, case 2: self-translated long Latin quotations.} For long
Latin passages \spire{} supplies its \emph{own} English rendering rather than an
established scholarly translation. On H004 it quotes Aulus Gellius's
\emph{Noctes Atticae} at length:
\begin{quote}\small
``Natura'' inquit ``omnium rerum, quae nos genuit, induit nobis inoleuitque
\ldots\ amorem nostri et caritatem\ldots'' (\emph{Noctes Atticae}) [E5]
\end{quote}
and renders it in its own prose (``nature instills in every creature \ldots\ an
innate love and care for itself''). Humanities citation norms expect the
canonical published translation (e.g.\ the Loeb Classical Library edition) for
such passages because they are accurate, vetted, and citable. Pairing the retrieved primary text with a
registered translation resource would resolve this.

\paragraph{GraphRAG: secondary literature substituted for the primary text
(severe).} On H008 (Daoism, Confucianism, and Augustine on learning), asked
about the \emph{Daoist} position, GraphRAG never reaches Laozi or Zhuangzi;
it instead retrieves a Neo-Confucian critic and a Qing philologist, and imports
Buddhism the question never raised:
\begin{quote}\small
The retrieved passage from Zhang Zai criticizes Buddhism and Daoism\ldots\ The passage
from Dai Zhen notes that Laozi said ``abandon learning and be free of worry''\ldots
\end{quote}
Expert~B scored EvidenceQuality~$=1$ and noted that the Daoist side used neither
Zhuangzi nor direct Daoist evidence, but secondary literature from Zhang Zai and even introduced Buddhism. Community-summary retrieval surfaces
commentary \emph{about} the topic rather than the cited primary source.

\paragraph{Text RAG: off-topic retrieval, then refusal (severe).} On CROSS\_006
(``should Confucius be regarded as a \emph{Sinicus Cicero}?''), flat cosine
retrieval returns only unrelated Latin rhetoric (Quintilian's \emph{Institutio
Oratoria}, Seneca's \emph{De Brevitate Vitae} and \emph{Epistulae}), with no material on
Confucius or the \emph{Confucius Sinarum Philosophus}, so the answer collapses
to a single sentence:
\begin{quote}\small
``The retrieved passages do not contain any information about Confucius \ldots
Therefore, they are insufficient to answer whether Confucius should be regarded
as a Sinicus Cicero.''
\end{quote}
When retrieval misses, the flat baseline neither supplements nor reasons; its
answers collapse to such refusals (median length ${\sim}270$ characters), which
the judge penalises on every axis.

\paragraph{Naive LLM: ungrounded recall (severe).} With no retrieval, the
no-retrieval baseline answers CROSS\_001 (Confucius vs.\ Cicero) fluently from
parametric memory and cites from recollection:
\begin{quote}\small
\{source: \emph{Lunyu / Analects} 12.1, ``to master oneself and return to ritual is \emph{ren}''; role: central;
confidence: high\}; \{source: \emph{De officiis}\ldots\}
\end{quote}
The named loci are plausible but unverifiable against the corpus and prone to
misattribution. This failure motivates evidence-grounded retrieval.

\subsection{Failure modes of generic agent baselines}
\label{app:agentic-failures}

PRG and ReAct use the same retrieval substrate and generator as \spire{}.
Their paper-level diagnostics therefore help locate the effect of the agent
workflow. Table~\ref{tab:agentic-failures} summarizes two distinct patterns.

\begin{table}[t]
\centering\scriptsize
\setlength{\tabcolsep}{3.2pt}
\begin{tabular}{@{}llrrr@{}}
\toprule
System & Diagnostic & Mean units & Overall & Cross \\
\midrule
PRG & Pool judged irrelevant & 50.0 & 34.2\% & 51.6\% \\
ReAct & No retrieved unit cited & 135.9 & 15.4\% & 21.6\% \\
ReAct & Gold work in pool, none named & 135.9 & 4.9\% & n/a \\
\bottomrule
\end{tabular}
\caption{Paper-level evidence-use diagnostics for the generic agents. Mean
units is the retrieved volume per paper.}
\label{tab:agentic-failures}
\end{table}

\paragraph{PRG: useful evidence judged irrelevant.} PRG retrieves a fixed
pool of $50$ units before synthesis. On CROSS\_006, which asks whether
Confucius should be regarded as a \emph{Sinicus Cicero}, that pool covers all
gold evidence in the reference paper, with $\mathrm{EvidenceR}=100.0$.
Its answer nevertheless concludes that the retrieved texts provide no
relevant evidence for the comparison. The missing step is alignment: the
Latin and Chinese passages must be treated as two sides of the same question.
This pattern appears in $51.6\%$ of cross-tradition papers, compared with
$24.3\%$ of single-work Chinese papers. It is the type of evidence decision
supported by the Comparing primitive.

\paragraph{ReAct: retrieval without consistent binding.} ReAct retrieves
$135.9$ units per paper on average. The larger pool expands coverage, yet
$15.4\%$ of its essays cite none of the retrieved units. In another $4.9\%$,
the pool contains a gold work that the essay never names. The zero-citation
rate rises to $21.6\%$ for cross-tradition questions, where evidence from both
sides must be aligned before synthesis. This pattern corresponds to the role
of Illustrating, which links claims to supporting passages.

The category results show the same progression at finer granularity. On
cross-tradition questions, PRG and ReAct reach EvidenceR scores of $17.7$ and
$21.4$, but their WorkR scores are $8.4$ and $10.6$, and their SectionR scores
are $3.4$ and $1.3$ (Table~\ref{tab:app-bycat}). Generic planning broadens the
pool. Scholarly primitives more consistently turn that pool into localized,
citable evidence.

The taxonomy therefore qualifies the aggregate eR result. \spire{} still has
true retrieval failures and scope errors, but the largest category is a
locator-level gap and some strict misses remain defensible alternative paths.
The baseline cases show a different concentration of failure: wrong
source layers, unusable retrieval, and unsupported from-memory citation. These
patterns explain why work/section recall and EvidenceQuality separate the
systems more clearly than surface similarity.

\section{Graph, Community and Cluster Substrate}
\label{app:graph}

\subsection{Intrinsic semantic-cluster evaluation}
\label{app:cluster-intrinsic}
\begin{figure*}[tbp]\centering
\includegraphics[width=\textwidth]{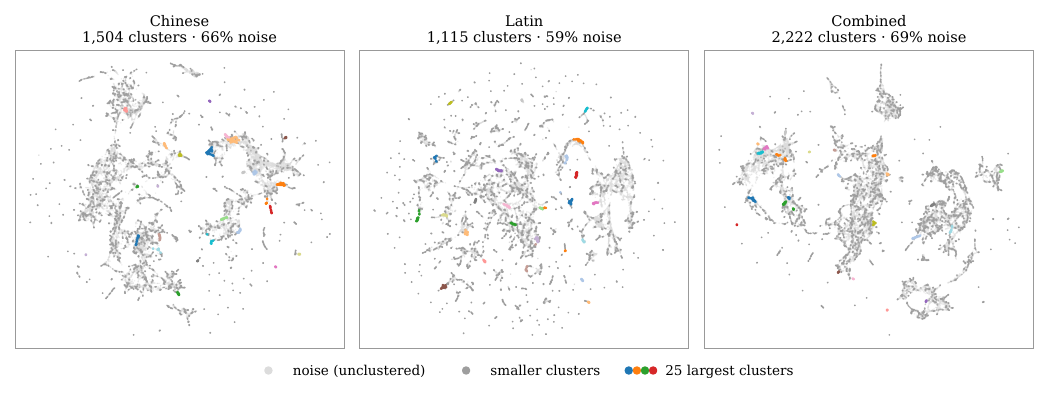}
\caption{UMAP projections of the HDBSCAN semantic-cluster spaces for the
Chinese, Latin, and combined cross-context indices at minimum cluster size 10
(left to right). Panel titles report cluster counts and noise shares. Colored
points mark the 25 largest clusters; dark gray points are smaller non-noise
clusters, and light gray points are HDBSCAN noise. UMAP is used only for
visualization; Table~\ref{tab:cluster-intrinsic} evaluates clusters in the
original $1{,}024$-dimensional embedding space.}
\label{fig:cluster-intrinsic}
\end{figure*}

We evaluate HDBSCAN in the original $1{,}024$-dimensional embedding space
against cluster-count-matched KMeans and random-label controls. HDBSCAN has the
highest silhouette at every evaluated scale (Table~\ref{tab:cluster-intrinsic}).
Figure~\ref{fig:cluster-intrinsic} visualizes the resulting cluster geometry.
As minimum cluster size decreases,
within-cluster cosine rises from $0.539$ to $0.664$ while between-cluster cosine
stays near $0.36$. This intrinsic separation complements the downstream result:
removing the cluster tier lowers eR from $44.3$ to $32.5$.

\begin{table*}[tbp]\centering\small\setlength{\tabcolsep}{8pt}
\begin{tabular}{@{}rrrrrrr@{}}
\toprule
Min. size & Coverage & HDBSCAN sil. & KMeans sil. & Random sil. & Intra cos & Inter cos \\
\midrule
100 & 24.4\% & 0.060 & 0.053 & $-0.060$ & 0.539 & 0.352 \\
30  & 26.1\% & 0.077 & 0.036 & $-0.140$ & 0.589 & 0.355 \\
10  & 30.4\% & 0.093 & $-0.051$ & $-0.241$ & 0.643 & 0.360 \\
3   & 41.4\% & 0.167 & n/a & $-0.262$ & 0.664 & 0.370 \\
\bottomrule
\end{tabular}
\caption{Intrinsic evaluation of semantic clusters. Coverage is the percentage
of embedded entities assigned to non-noise clusters.}
\label{tab:cluster-intrinsic}
\end{table*}

\subsection{Graph and cluster artefacts per scale}
Table~\ref{tab:artifacts} confirms the design rationale of \S\ref{sec:retrieval}: even on the combined \texttt{cross} corpus the mean cross-language balance of the coarsest, most reliable clusters is $0.008$. The vector index captures cross-context recurrence; the Comparing primitive performs cross-tradition alignment at the agent level.
\begin{table*}[t]
\centering
\scriptsize
\setlength{\tabcolsep}{3pt}
\renewcommand{\arraystretch}{0.9}
\begin{tabular}{@{}lllrrrr@{}}
\toprule
Lang & Algo & Scale & \#groups & max & avg sz & x-lang bal \\
\midrule
zh & Leiden & res=2.0 & 1,330 & 6,218 & 81.0 & -- \\
zh & Leiden & res=8.0 & 1,562 & 3,451 & 69.0 & -- \\
zh & Leiden & res=20.0 & 2,072 & 2,356 & 52.0 & -- \\
zh & Leiden & res=30.0 & 2,488 & 2,290 & 43.3 & -- \\
zh & HDBSCAN & mcs=100 & 116 & 1,684 & 315.9 & 0.000 \\
zh & HDBSCAN & mcs=30 & 422 & 629 & 91.9 & 0.000 \\
zh & HDBSCAN & mcs=10 & 1,504 & 395 & 29.4 & 0.000 \\
zh & HDBSCAN & mcs=3 & 7,753 & 59 & 7.4 & 0.000 \\
la & Leiden & res=2.0 & 776 & 5,129 & 80.4 & -- \\
la & Leiden & res=8.0 & 949 & 1,456 & 65.7 & -- \\
la & Leiden & res=20.0 & 1,319 & 1,199 & 47.3 & -- \\
la & Leiden & res=30.0 & 1,603 & 1,139 & 38.9 & -- \\
la & HDBSCAN & mcs=100 & 68 & 1,663 & 337.5 & 0.000 \\
la & HDBSCAN & mcs=30 & 291 & 607 & 89.9 & 0.000 \\
la & HDBSCAN & mcs=10 & 1,115 & 211 & 27.7 & 0.000 \\
la & HDBSCAN & mcs=3 & 5,006 & 45 & 7.6 & 0.000 \\
cross & HDBSCAN & mcs=100 & 147 & 2,472 & 346.5 & 0.008 \\
cross & HDBSCAN & mcs=30 & 563 & 854 & 96.8 & 0.020 \\
cross & HDBSCAN & mcs=10 & 2,222 & 367 & 28.6 & 0.059 \\
cross & HDBSCAN & mcs=3 & 11,958 & 66 & 7.2 & 0.169 \\
\bottomrule
\end{tabular}
\caption{Offline graph/cluster artifacts per scale. ``x-lang bal'' is the
mean cross-language balance of HDBSCAN clusters; it is near zero even on the
combined \texttt{cross} corpus, empirically justifying agent-level
cross-tradition alignment (\S\ref{sec:retrieval}).}
\label{tab:artifacts}
\end{table*}

\section{Benchmark Record and Extracted-Graph Examples}
\label{app:examples}

\lstdefinestyle{examplepromptstyle}{
  basicstyle=\ttfamily\tiny,
  breaklines=true,
  breakindent=0pt,
  columns=fullflexible,
  keepspaces=true,
  numbers=left,
  numberstyle=\tiny\color{gray},
  numbersep=5pt,
  showstringspaces=false,
  upquote=true,
}
\newtcblisting{examplepromptbox}[1]{
  breakable, enhanced, listing only,
  listing options={style=examplepromptstyle},
  colback=gray!7, colframe=gray!55, boxrule=0.35pt,
  arc=1pt, left=1.7em, top=1pt, bottom=1pt,
  fonttitle=\bfseries\scriptsize, coltitle=black,
  colbacktitle=gray!18, title={#1}}

\paragraph{Benchmark record.} A representative gold record (one of $406$;
\S\ref{sec:benchmark}). Internal identifiers and annotation labels are
omitted; we keep the fields a system is scored against.

\begin{examplepromptbox}{Gold benchmark record (abridged)}
paper: "Pietas in pro Sexto Roscio of Cicero and Confucian xiao"
paper_language: en   literature_languages: zh_classical + la_classical

research_question:
  How does Cicero's pietas in pro Sexto Roscio relate to Confucian xiao,
  and what does comparison show about filial duty and social order?

research_findings (excerpt):
  Parricide is treated as the ultimate filial impiety; Cicero rejects the
  charge by arguing ex persona, ex causa, and ex facto ipso.

evidences (2 of 13 shown):
  E01  Pro Roscio Amerino  loc 37  translated_quote  central
       "Sextus Roscius stands accused of the murder of his father. ..."
  E04  Pro Roscio Amerino  loc 69  original_quote    supporting
       "quod in impios singulare supplicium invenerunt"
\end{examplepromptbox}

\paragraph{Extracted graph.} Selected fields from per-chunk
\textsc{DeepSeek-V4-Flash} output (\S\ref{sec:retrieval}) are shown below.
The arrays are shortened after schema validation; Chinese glyphs are romanized
to match the paper's convention.

\begin{examplepromptbox}{Extracted graph -- Latin chunk (Ammianus Marcellinus, Res Gestae)}
{"entities":[
  {"type":"Person","title_modern":"Ammianus Marcellinus",
   "description_modern":"Roman historian and author of the Res Gestae.",
   "evidence_spans":[1,9]},
  {"type":"Person","title_modern":"Nohodares",
   "description_modern":"A Persian nobleman who planned an invasion of Mesopotamia.",
   "evidence_spans":[2]},
  {"type":"Concept","domain":"Politics & Governance","title_modern":"Saracen customs",
   "description_original":"Saracenorum irruptiones et mores",
   "description_modern":"The nomadic, predatory life and military habits of the Saracens.",
   "evidence_spans":[9,10,11,12,13,14,15]},
  {"type":"Concept","title_modern":"Persian stratagem",
   "description_original":"Persarum commentum irritum",
   "description_modern":"A failed Persian military plan or ruse.","evidence_spans":[1]}],
 "relations":[
  {"source":"Ammianus Marcellinus","target":"Saracen customs","relation_name":"describes",
   "description":"Ammianus gives an ethnographic account of Saracen customs.",
   "evidence_spans":[9,...,15]},
  {"source":"Nohodares","target":"Persian stratagem","relation_name":"devised",
   "description":"Nohodares planned the invasion of Batnae during its annual fair.",
   "evidence_spans":[2,...,7]}]}
\end{examplepromptbox}

\begin{examplepromptbox}{Extracted graph -- Chinese chunk (Liu Mi, Sanjiao Pingxin Lun; romanized)}
{"entities":[
  {"type":"Person","title_modern":"Liu Mi",
   "description_modern":"Author of the Sanjiao Pingxin Lun.",
   "evidence_spans":[25,66]},
  {"type":"Concept","domain":"Religion & Spirituality","title_modern":"fangbian (skillful means)",
   "description_modern":"A Buddhist term for expedient teaching; here the expedient by which
    earlier worthies reconciled the Three Teachings.","evidence_spans":[2,19,20,28]},
  {"type":"Concept","title_modern":"sanjiao (Three Teachings)",
   "description_modern":"Confucianism, Buddhism, and Daoism; the text argues they can be unified.",
   "evidence_spans":[25,34]},
  {"type":"Person","title_modern":"Cheng Yi",
   "description_modern":"Northern-Song Neo-Confucian; here decries Buddhism as heterodox and monstrous.",
   "evidence_spans":[59,90]}],
 "relations":[
  {"source":"Liu Mi","target":"sanjiao","relation_name":"argues_for_unity",
   "description":"Liu Mi holds the Three Teachings can be merged into one.","evidence_spans":[25,26,27]},
  {"source":"Liu Mi","target":"Cheng Yi","relation_name":"refutes",
   "description":"Liu Mi rebuts Cheng Yi, noting the Confucian classics also contain strange tales.",
   "evidence_spans":[66,...,92]}]}
\end{examplepromptbox}

\subsection{Corpus access and licensing}
\label{app:artifacts}

The source texts are under copyright and cannot be redistributed. The
repository therefore ships catalogues for obtaining legally licensed
copies together with derived, non-infringing artefacts. The Chinese corpus
is built from \emph{Zhongguo Xueshu Mingzhu Tiyao} (\emph{Compendium of
Chinese Academic Masterworks}, Fudan University Press)
\citep{zhongguo1992mingzhu}; the Latin corpus is built from \emph{The Latin
Library} \citep{latinlibrary}. For both, we release the work list and
extracted graph and cluster data in JSONL, but no raw full text. For the
peer-reviewed-paper benchmark, we provide the paper list and extracted
research questions, findings, and cited primary-source evidence in JSON
rather than copyrighted PDFs. Scripts reconstruct restricted components
when redistribution is not allowed. Released data are limited to scholarly
metadata, structured annotations, and derived representations needed for
reproducibility.



\section{Prompt Contracts and Output Schemas}
\label{app:agentprompts}
The released repository contains the literal system prompts, machine-readable
schemas, and worked Chinese and Latin examples. This section records the
contracts that determine each stage and the validation rules used in the
experiments.

\subsection{Agent workflow contracts}

\begin{description}[style=nextline,leftmargin=1em,
  itemsep=1pt,parsep=0pt,topsep=2pt]
\item[Discovering.] A research question becomes two to four sub-questions with
languages, literal anchors, and retrieval granularity.
\item[Annotating.] Passages return question-directed interpretations, links, and
relevance in JSON keyed by source unit.
\item[Comparing.] The two labeled sides of the shared EvidencePool return
parallels, divergences, unilateral concepts, and inverted emphases. Analogues
are treated as interpretive proposals, and both sides must supply evidence.
\item[Referring.] Provenance checks produce follow-up queries naming a work,
author, concept, or native term rather than a workflow instruction.
\item[Sampling.] The EvidencePool yields $8$--$15$ passage IDs with checks for
question, source, and tradition coverage.
\item[Illustrating.] Claims are exemplified with retrieved passages, and each
verbatim quotation is bound to a stable \texttt{[Eid]}.
\item[Representing.] The shared synthesis call writes the essay, then emits a
structured audit that preserves the same evidence IDs. Each audited quotation
resolves to a retrieved source unit.
\item[Reflection.] Coverage gaps and weakly supported claims may trigger up to
two new retrieval rounds. Each round is tied to a diagnosed gap.
\end{description}

\subsection{Graph, report, and audit schemas}
\label{app:prompts}
Graph extraction returns \texttt{entities} and \texttt{relations}. Every entity
contains a type from Concept, Person, or Work, a canonical title, a short
description, and sentence-index evidence spans. Each directed relation names
two extracted entities, their types, a relation label, a grounded description,
and its evidence spans. Relation endpoints must resolve to entities in the same
validated output.

Community and semantic-cluster reports share one schema. It contains a title,
summary, narrative, key findings, top entities, themes, a rating, and a rating
explanation. Community reports summarize supplied graph relations. Cluster
reports describe semantic affinity and mark it as distinct from documented
textual connection. This distinction prevents embedding proximity from being
reported as historical influence.

All structured stages pass JSON-schema validation. Source identifiers remain
stable across retrieval, annotation, curation, prose citation, and the final
audit. Cross-tradition questions require evidence from both source pools. The
final audit records findings, evidence IDs, source coverage, and targeted
follow-up queries. Ablations remove the corresponding stage while preserving
the contracts of all remaining stages.

\section{Scholarly-Judge Rubric and Prompt}
\label{app:judge}
Both LLM raters and the human experts use one rubric, given verbatim
below. In the blind comparative protocol the rater sees all six
systems' anonymised essays for one question in random order and scores
each independently on the same scale; it never sees the paper's evidence
in the \emph{free} track.

\begin{promptbox}{Scholarly-judge system prompt (humanities, comparative)}
You are a humanities scholar reviewing research-assistance material (not a publication-ready
paper). You are given the research question, the paper title that produced it (background only,
NOT a thesis to replicate), and the essay under review. In the comparative protocol you see
SEVERAL anonymised candidate essays for the SAME question in random order; score EACH
independently on the same 1-5 rubric. Length is not rewarded.

1. AnswerAccuracy -- does the essay precisely address the relation the
   question asks (how / why / in what sense / through what mechanism)
   with textually sound claims, not merely list related concepts?
   1 = off-topic / contradicts the texts; 2 = misreads the core
   concept; 3 = relevant but generic or half-answers the key relation
   (topical adequacy CAPS HERE); 4 = accurate, minor simplification;
   5 = precise on the core relation, comparanda kept distinct.

2. ArgumentDepth -- is the argument driven by specific textual detail
   and developed in layers, not paraphrastic summary?
   1 = no real argument; 2 = assertion/paraphrase only; 3 = a thesis
   but mainly generalisation or restatement (SUMMARY CAPS HERE);
   4 = structured multi-step argument, several moves text-driven;
   5 = layered analysis deriving claims from close reading of specific
   passages.

3. CoverageCompleteness -- does it cover the sub-tasks the question
   entails? (A comparison requires both sides + similarity + difference
   + the limits of the comparison.)
   1 = misses major dimensions; 2 = one side only; 3 = main question
   answered but a key sub-dimension thin (a single general point CAPS
   HERE); 4 = main dimensions covered, minor omissions; 5 = thorough
   and handles nuance / limitations.

4. EvidenceQuality -- how well is the answer grounded in specific
   primary-text evidence: density of direct quotation, specificity of
   citation, close reading of the source material, versus vague
   work-name dropping or hand-waving paraphrase? (Faithfulness floor:
   treat a citation as fabricated only if the quoted words clearly do
   not belong to the cited work or are mis-attributed; this is rare and
   not the main concern.)
   1 = essentially no primary-text evidence or clear fabrication;
   2 = evidence very sparse, loosely connected; 3 = some real evidence
   but sparse/generic or bare work-name mention (~no direct quotation
   CAPS HERE); 4 = key claims backed by specific, accurate quotation;
   5 = quotation specific and tightly integrated with close-reading
   analysis throughout.

Output VALID JSON only. Single essay:
{"AnswerAccuracy": <1-5>, "ArgumentDepth": <1-5>,
 "CoverageCompleteness": <1-5>, "EvidenceQuality": <1-5>,
 "rationale": "<one short sentence per axis, in order>"}
Comparative (N essays):
{"scores": {"Answer 1": {"AnswerAccuracy": <1-5>, ...},
 "Answer 2": {...}, ...},
 "comparative_note": "<which answer is most grounded in direct
 primary-text evidence, which is mostly generic summary>"}
\end{promptbox}

\section{Fragment-Level Qualitative Comparisons}
\label{app:qual-fragment-figures}

\textbf{Reading guide.} The examples compare representative fragments from
the same question. Green marks direct evidence, cyan marks evidence-to-claim
reasoning, blue marks broad theses, orange marks source locators, purple marks
graph summaries, and red marks limitations. Ellipses omit surrounding text.

\newtcolorbox{qualquestion}{
  enhanced, colback=gray!4, colframe=gray!55, boxrule=0.45pt,
  arc=1.5pt, left=5pt, right=5pt, top=4pt, bottom=4pt,
  before skip=4pt, after skip=6pt}

\newtcolorbox{qualnaive}[1]{
  enhanced, colback=blue!4, colframe=blue!55!black, boxrule=0.45pt,
  arc=1.5pt, left=5pt, right=5pt, top=3pt, bottom=4pt,
  colbacktitle=blue!13, coltitle=black, fonttitle=\bfseries\footnotesize,
  title={#1}, before skip=3pt, after skip=3pt}

\newtcolorbox{qualtext}[1]{
  enhanced, colback=orange!5, colframe=orange!75!black, boxrule=0.45pt,
  arc=1.5pt, left=5pt, right=5pt, top=3pt, bottom=4pt,
  colbacktitle=orange!14, coltitle=black, fonttitle=\bfseries\footnotesize,
  title={#1}, before skip=3pt, after skip=3pt}

\newtcolorbox{qualgraph}[1]{
  enhanced, colback=violet!5, colframe=violet!65!black, boxrule=0.45pt,
  arc=1.5pt, left=5pt, right=5pt, top=3pt, bottom=4pt,
  colbacktitle=violet!13, coltitle=black, fonttitle=\bfseries\footnotesize,
  title={#1}, before skip=3pt, after skip=3pt}

\newtcolorbox{qualspire}[1]{
  enhanced, colback=green!5, colframe=green!55!black, boxrule=0.55pt,
  arc=1.5pt, left=5pt, right=5pt, top=3pt, bottom=4pt,
  colbacktitle=green!14, coltitle=black, fonttitle=\bfseries\footnotesize,
  title={#1}, before skip=3pt, after skip=3pt}

\newtcolorbox{qualprg}[1]{
  enhanced, colback=brown!4, colframe=brown!55!black, boxrule=0.45pt,
  arc=1.5pt, left=5pt, right=5pt, top=3pt, bottom=4pt,
  colbacktitle=brown!13, coltitle=black, fonttitle=\bfseries\footnotesize,
  title={#1}, before skip=3pt, after skip=3pt}

\newtcolorbox{qualreact}[1]{
  enhanced, colback=cyan!5, colframe=cyan!55!black, boxrule=0.45pt,
  arc=1.5pt, left=5pt, right=5pt, top=3pt, bottom=4pt,
  colbacktitle=cyan!14, coltitle=black, fonttitle=\bfseries\footnotesize,
  title={#1}, before skip=3pt, after skip=3pt}

\newcommand{\qualfragment}[2]{%
  \Needspace{7\baselineskip}%
  \medskip
  \noindent\textbf{#1}\par
  \noindent\textit{Dimension: #2}\par\smallskip}

\newcommand{\qualchineseblock}[1]{%
  \noindent\includegraphics[width=0.96\columnwidth]{#1}\par\smallskip}

\subsection{Chinese Case Study}

\begin{qualquestion}
\includegraphics[width=\linewidth]{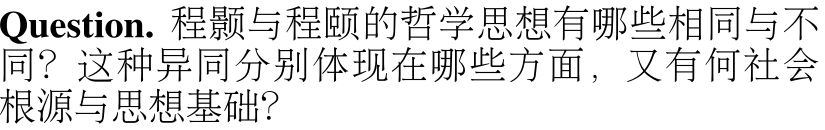}
\end{qualquestion}

\qualfragment{Fragment 1: Core distinction}{argument specificity}

\qualchineseblock{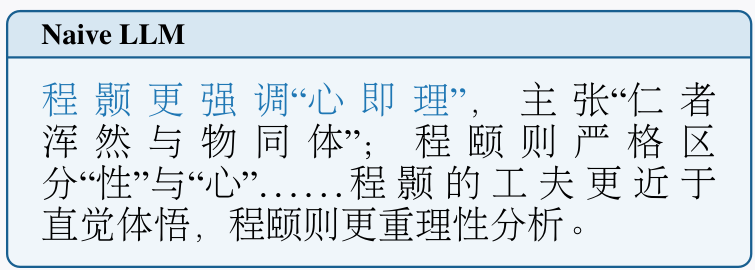}
\qualchineseblock{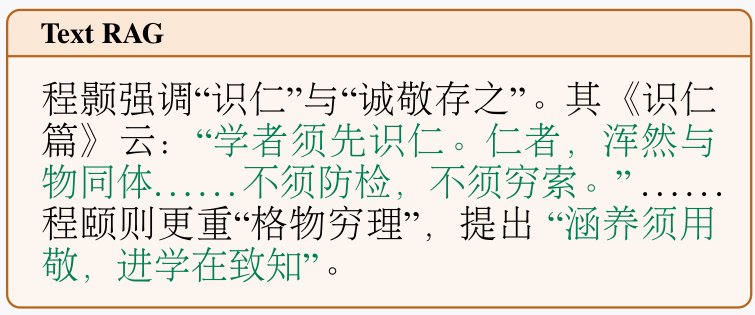}
\qualchineseblock{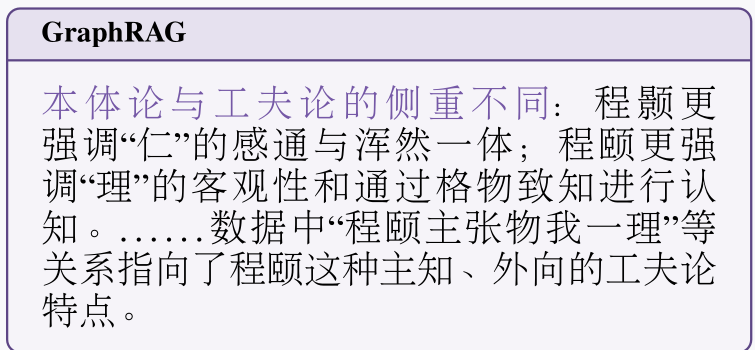}
\qualchineseblock{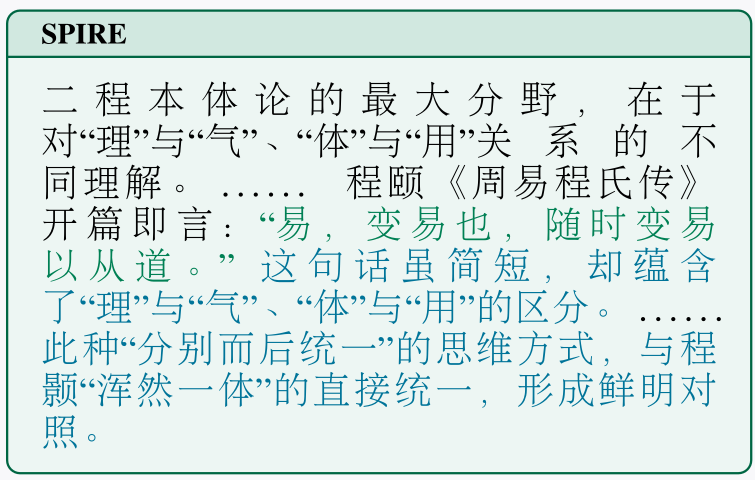}

\noindent\textit{Analysis.} The Naive LLM states a plausible textbook
contrast, Text RAG adds relevant quotations, and GraphRAG broadens the
conceptual map. \spire{} turns a textual phrase into an interpretive mechanism:
the contrast is derived from how the two thinkers handle \emph{li}, \emph{qi},
\emph{ti}, and \emph{yong}.

\qualfragment{Fragment 2: Evidence use}{evidence-to-claim reasoning}

\qualchineseblock{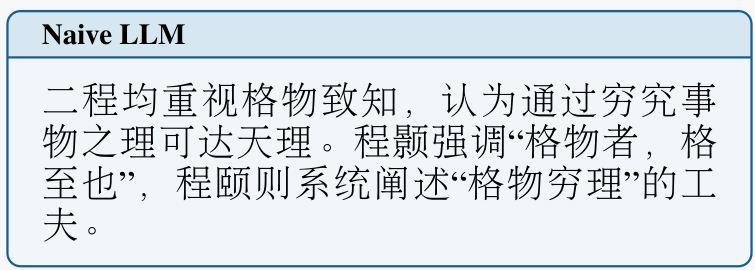}
\qualchineseblock{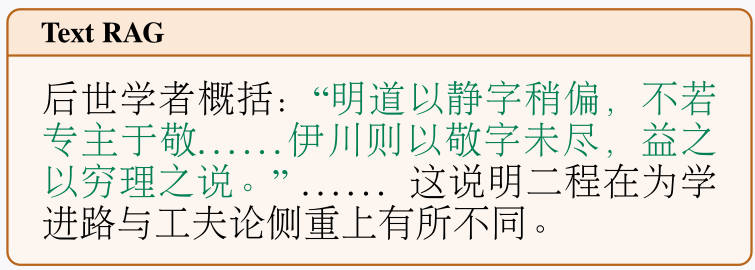}
\qualchineseblock{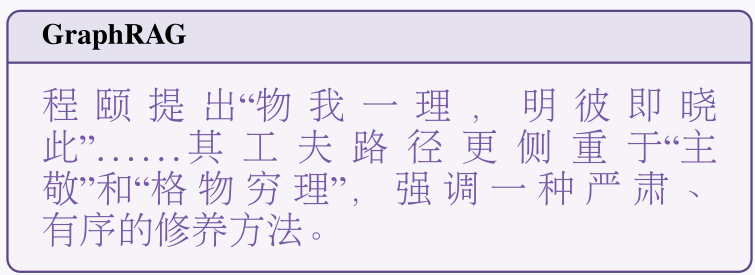}
\qualchineseblock{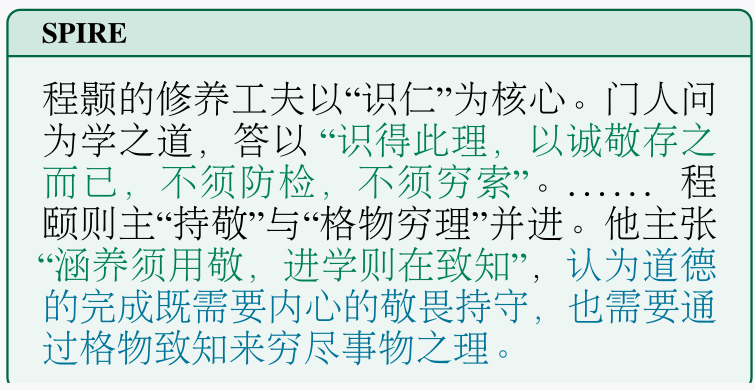}

\noindent\textit{Analysis.} The systems differ in how they use citations.
\spire{} uses paired quotations to construct a comparative argument about
cultivation. Cheng Hao's method is preservative and non-searching, while Cheng
Yi combines reverent self-cultivation with inquiry into principle.

\subsection{Latin Case Study}

\begin{qualquestion}
\textbf{Question.} How does Cicero's \emph{De Divinatione} compare evidence
and explanation in assessing divination, and what does Marcus' critique really
depend on?
\end{qualquestion}

\qualfragment{Fragment 1: What Marcus' critique depends on}{claim precision}

\begin{qualnaive}{Naive LLM}
Marcus' critique depends on a distinction between correlation and causation:
even if some predictions come true, this does not prove a causal link between
the sign and the event. Marcus employs a form of skeptical reasoning. \ldots
\end{qualnaive}

\begin{qualtext}{Text RAG}
Marcus' critique, as presented in Book II, does not directly appear in the
provided passages. \ldots{} The passages show that Quintus' case is built on
empirical observation and metaphysical commitment. Marcus' likely strategy is
to challenge the reliability of the evidence and the coherence of the
explanation.
\end{qualtext}

\begin{qualgraph}{GraphRAG}
Marcus' critique ultimately depends not on a dogmatic rejection of the gods,
but on a rationalist and skeptical demand for a coherent, logical, and
non-superstitious basis for belief.
\end{qualgraph}

\begin{qualspire}{SPIRE}
Marcus' critique depends, in the final analysis, on a rigorous distinction
between correlation and causation that Quintus systematically blurs. When
Quintus presents the case of Crassus' disaster, \ldots{} he insists that the
signs \emph{non causas adferunt, cur quid eveniat, sed nuntiant eventura}
(I.30) [E5]. The distinction is revealing: Quintus himself admits that signs do
not cause events, they only announce them.
\end{qualspire}

\noindent\textit{Analysis.} All four systems identify the same general issue,
but they differ in how strongly the claim is grounded. \spire{} reconstructs
the argument from a specific Latin sentence. It shows that the distinction
between correlation and causation arises from Quintus' own formulation.

\qualfragment{Fragment 2: Evidence versus explanation}{close reading}

\begin{qualnaive}{Naive LLM}
Quintus marshals historical examples, dreams, oracles, and portents. \ldots{}
Marcus argues that alleged successes are either coincidental, exaggerated, or
self-fulfilling prophecies.
\end{qualnaive}

\begin{qualtext}{Text RAG}
Quintus opens by asserting that divination is \emph{the foretelling and
foreknowledge of things that are thought to be fortuitous}. \ldots{} He
compares the unfolding of time to the unwinding of a rope. \ldots{} The
evidence gaps are significant: we lack Marcus' actual counterarguments.
\end{qualtext}

\begin{qualgraph}{GraphRAG}
The critique demands a higher standard of evidence. Cicero contrasts the
unreliable predictions of divination with reliable knowledge derived from
other human arts, like medicine. \ldots{} Failures and ambiguities outweigh
occasional successes.
\end{qualgraph}

\begin{qualspire}{SPIRE}
Quintus observes that the ancients accepted divination \emph{rerum magis
eventis moniti quam ratione docti} (I.5) [E1], warned by outcomes rather than
taught by reason. The phrase captures the tension that will dominate the
entire debate: events provide the raw material, but reason must decide whether
those events warrant the conclusions drawn from them.
\end{qualspire}

\noindent\textit{Analysis.} The baselines discuss evidence and explanation
mostly as abstract categories. \spire{} identifies a compact Latin phrase and
turns it into the organizing problem of the dialogue.

\qualfragment{Fragment 3: Reconstructing the positive case}{balanced coverage}

\begin{qualnaive}{Naive LLM}
Quintus relies on Stoic \emph{sympatheia} and divine providence as explanatory
frameworks, claiming that the gods communicate through signs that trained
interpreters can read.
\end{qualnaive}

\begin{qualtext}{Text RAG}
Quintus' explanation ties divination to a rational, causal framework: if all
things happen by fate, and if the chain of causes can be grasped, then future
events are in principle knowable through their signs.
\end{qualtext}

\begin{qualgraph}{GraphRAG}
On one hand, Cicero argues for the validity of divination, grounding it in the
existence of gods and their providential care for humanity, supported by
logical arguments, appeals to fate and nature, and extensive historical
examples.
\end{qualgraph}

\begin{qualspire}{SPIRE}
Quintus writes: \emph{Similis est haruspicum responsio omnisque opinabilis
divinatio; coniectura enim nititur. \ldots{}} (II.24--25) [E2]. The logic is
inductive: repeated observation of correlations between antecedent signs and
subsequent events has produced an art. This comparison is crucial because it
frames divination as a species of empirical knowledge. It is \emph{opinabilis},
a matter of conjecture, but conjecture disciplined by experience.
\end{qualspire}

\noindent\textit{Analysis.} \spire{} first reconstructs Quintus' strongest
case: divination as an empirical and conjectural art. This balanced account
sharpens the later critique.

\subsection{Agentic Baselines on a Matched Question}
\label{app:qual-agentic}

\begin{qualquestion}
\textbf{Question.} How do Mencius and Xunzi differ in their understanding of
\emph{li} as a virtue, and what follows for the relation between moral and
civic virtue?
\end{qualquestion}

This example comes from GLOBAL\_ZH\_007. Both agents retrieve from the shared
store, so the comparison focuses on how they turn retrieved material into an
answer.

\begin{qualprg}{PRG}
For Mencius, \emph{li} is an internal virtue rooted in the innate goodness of
human nature, expressed through the four sprouts. \ldots{} For Xunzi,
\emph{li} is an external instrument created by sages to restrain and reform
human nature. \ldots{} The result is an integrative model in Mencius, where
moral and civic virtue are continuous and internally grounded, and a
constructivist model in Xunzi, where civic virtue precedes and shapes moral
virtue through external ritual discipline.
\end{qualprg}

\begin{qualreact}{ReAct}
As Mencius states: The heart of compassion is the sprout, \emph{duan}, of
\emph{ren} \ldots{} the heart of reverence and respect, \emph{gongjing zhi
xin}, is the sprout of \emph{li} \ldots{} I originally possess them
(\emph{Mencius} 6A6).
\end{qualreact}

\noindent\textit{Analysis.} PRG gives a clear and defensible contrast without
a quotation or textual locator. ReAct retains the source terms and cites a
specific passage, making its evidence use easier to inspect. Across the five
categories, both agents remain below \spire{} in WorkR and SectionR
(Table~\ref{tab:app-bycat}). The gap reflects the conversion of retrieved
material into claims bound to specific evidence.

\end{document}